\documentclass[lettersize,journal]{IEEEtran}
\usepackage{amsmath,amsfonts}
\usepackage{algorithmic}
\usepackage{algorithm}
\usepackage{array}
\usepackage[caption=false,font=normalsize,labelfont=sf,textfont=sf]{subfig}
\usepackage{textcomp}
\usepackage{stfloats}
\usepackage{url}
\usepackage{verbatim}
\usepackage{graphicx}
\usepackage{cite}

\usepackage{booktabs}
\usepackage{multirow}
\usepackage{amssymb}

\usepackage{newfloat}
\usepackage{listings}

\usepackage{tabularx}
\usepackage{hyperref}

\newcommand{\ie}{\textit{i.e.},\ }
\newcommand{\eg}{\textit{e.g.},\ }

\begin{document}

\title{RoDE: Linear Rectified Mixture of Diverse Experts for Food Large Multi-Modal Models}

\author{Pengkun~Jiao,
        Xinlan~Wu,
        Bin~Zhu,
        Jingjing~Chen,
        Chong-Wah~Ngo and Yu-Gang~Jiang % <-this % stops a space
        
% \thanks{This paper was produced by the IEEE Publication Technology Group. They are in Piscataway, NJ.}% <-this % stops a space
% \thanks{Manuscript received April 19, 2021; revised August 16, 2021.}
\thanks{Pengkun Jiao is with the School of Computer Science, Fudan University. 
Shanghai, China. E-mail: pkjiao21@m.fudan.edu.cn. }
\thanks{Xinlan Wu is with the School of Computer Science, Fudan University. 
Shanghai, China. E-mail: 23210240332@m.fudan.edu.cn. }
\thanks{Bin Zhu is with the School of Computing and Information Systems, Singapore Management University. 
Singapore. E-mail: binzhu@smu.edu.sg. }
\thanks{Jingjing Chen is with the School of Computer Science, Fudan University. 
Shanghai, China. E-mail: chenjingjing@fudan.edu.cn. }
\thanks{Chong-Wah Ngo is with the School of Computing and Information Systems, Singapore Management University.
Singapore. E-mail:  cwngo@smu.edu.sg.}
\thanks{Yu-gang Jiang is with the School of Computer Science, Fudan University. 
Shanghai, China. E-mail: ygj@fudan.edu.cn. }
}
% The paper headers
\markboth{Journal of \LaTeX\ Class Files,~Vol.~14, No.~8, August~2021}%
{Shell \MakeLowercase{\textit{et al.}}: A Sample Article Using IEEEtran.cls for IEEE Journals}

% \IEEEpubid{0000--0000/00\$00.00~\copyright~2021 IEEE}
% Remember, if you use this you must call \IEEEpubidadjcol in the second
% column for its text to clear the IEEEpubid mark.

\maketitle

\begin{abstract}
Large Multi-modal Models (LMMs) have significantly advanced a variety of vision-language tasks. The scalability and availability of high-quality training data play a pivotal role in the success of LMMs. In the realm of food, while comprehensive food datasets such as Recipe1M offer an abundance of ingredient and recipe information, they often fall short of providing ample data for nutritional analysis. The Recipe1M+ dataset, despite offering a subset for nutritional evaluation, is limited in the scale and accuracy of nutrition information. To bridge this gap, we introduce Uni-Food, a unified food dataset that comprises over 100,000 images with various food labels, including categories, ingredients, recipes, and ingredient-level nutritional information. To mitigate the conflicts arising from multi-task supervision during fine-tuning of LMMs, we introduce a novel Linear \textbf{R}ectification Mixture \textbf{o}f \textbf{D}iverse \textbf{E}xperts (RoDE) approach. RoDE utilizes a diverse array of experts to address tasks of varying complexity, thereby facilitating the coordination of trainable parameters, \ie it allocates more parameters for more complex tasks and, conversely, fewer parameters for simpler tasks. RoDE implements linear rectification union to refine the router's functionality, thereby enhancing the efficiency of sparse task allocation. These design choices endow RoDE with features that ensure GPU memory efficiency and ease of optimization. Our experimental results validate the effectiveness of our proposed approach in addressing the inherent challenges of food-related multitasking. Project page at: \href{https://pengkun-jiao.github.io/UniFood-project/}{https://pengkun-jiao.github.io/UniFood-project/}.
\end{abstract}
%%
%% The code below is generated by the tool at http://dl.acm.org/ccs.cfm.
%% Please copy and paste the code instead of the example below.
%%
% \begin{CCSXML}
% <ccs2012>
% <concept>
% <concept_id>10002951.10003227.10003251.10003256</concept_id>
% <concept_desc>Information systems~Multimedia content creation</concept_desc>
% <concept_significance>500</concept_significance>
% </concept>
% </ccs2012>
% \end{CCSXML}

% \ccsdesc[500]{Information systems~Multimedia content creation}

%%
%% Keywords. The author(s) should pick words that accurately describe
%% the work being presented. Separate the keywords with commas.
% \keywords{Food classification, ingredient recognition, recipe generation, nutrition estimation, spare experts, LMMs}
%% A "teaser" image appears between the author and affiliation
%% information and the body of the document, and typically spans the
%% page.
% \begin{teaserfigure}
%   \includegraphics[width=\textwidth]{sampleteaser}
%   \caption{Seattle Mariners at Spring Training, 2010.}
%   \Description{Enjoying the baseball game from the third-base
%   seats. Ichiro Suzuki preparing to bat.}
%   \label{fig:teaser}
% \end{teaserfigure}

% \received{20 February 2007}
% \received[revised]{12 March 2009}
% \received[accepted]{5 June 2009}

%%
%% This command processes the author and affiliation and title
%% information and builds the first part of the formatted document.

\begin{figure}[t]
\centering
\includegraphics[width=\linewidth]{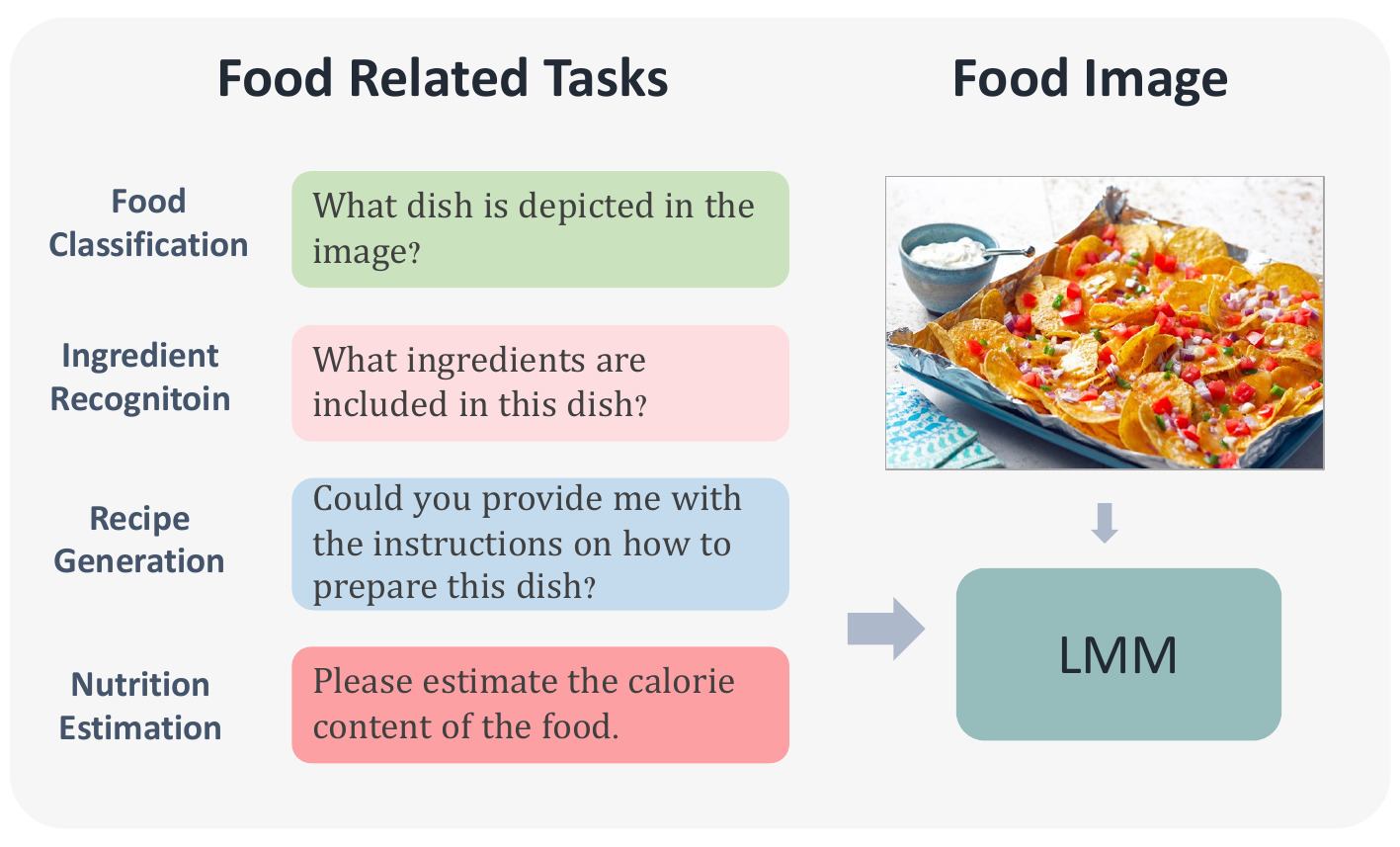}
\caption{RoDE's Emphasis on Food-Related VQA Tasks. RoDE primarily targets multi-task learning specific to food, \ie food classification, ingredient recognition, and nutrition estimation.
}
\label{img:teaser}
\end{figure}

\begin{figure*}[t]
\centering
\includegraphics[width=1\linewidth]{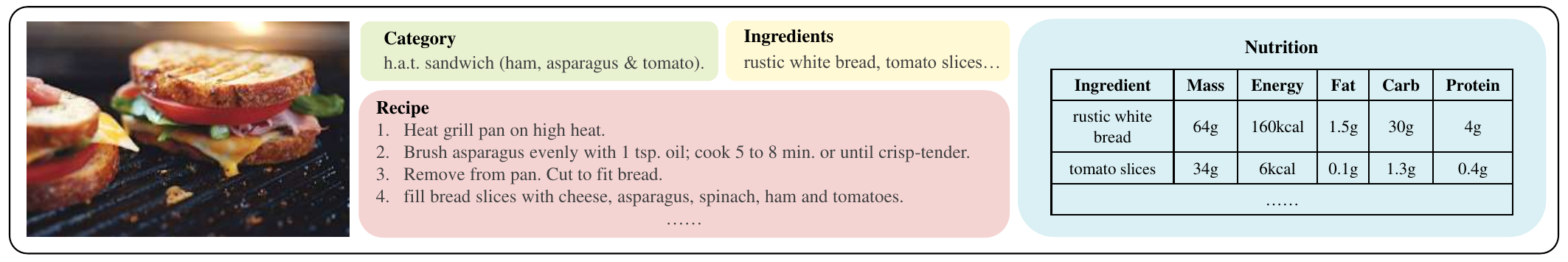}
\caption{\textbf{Composition of Labels in the UniFood Dataset.} The annotation consists of category, ingredient list, recipe instructions, and nutrition information.
}
\label{img:sample}
\end{figure*}

\section{Introduction}
Food occupies a central position in our daily lives, leading to the emergence of various food-related tasks, \eg ingredient recognition, recipe generation, and nutritional estimation~\cite{min2019survey}. These tasks have attracted considerable research interests over the years~\cite{chen2017cross_modal_recipe,jiang2019multi_scale_multi_view_feat_align,zhu2019r2gan,pan2020mm_cooking_workflow,chen2020mult_task_region_ingredient,zhu2021web_recipe_image_pairs,zhu2020cookgan,wang2022ingredient_guided_region_discovery,luo2023caclnet,liu2024from_canteen_2_daily,zhou2024synthesizing_knowledge_enhanced_features,liu2024convolution_enhanced_bibranch, song2024DAR}. 
Building on the success of Large Language Models (LLMs) \cite{radford2018bert, touvron2023llama2}, Large Multi-modal Models \cite{liu2023llava, luo2023caclnet} have begun to make a significant impact in many specialized areas, including the food domain \cite{yin2023foodlmm}.

The success of LLMs and LMMs can be largely attributed to the availability of large-scale training data. 
In the food domain,
while Recipe1M \cite{salvador2017recip1m} provides over a million web-crawled recipes and thousands of food images, it lacks comprehensive nutritional information. 
While Recipe1M+~\cite{marin2019recipe1m} includes a subset containing nutritional information, it suffers from limitations in data scale and annotation quality. 
Similarly, nutrition-focused datasets such as Nutrition5k~\cite{thames2021nutrition5k} are also limited in scale and may suffer from domain shift issues, as the labeled data primarily consists of lightly processed ingredients rather than fully cooked meals. Moreover, previous studies \cite{zhou2022dense_is_better, chen2024llavamole} have indicated that integrating diverse sources of data for training can lead to data conflicts. The performance of an LMM on a specific task also depends heavily on the representation of that task's data within the overall training dataset \cite{chen2024llavamole}. These challenges highlight the need for a unified dataset covering all food-related tasks from the same source, aiming at mitigating data conflicts and ensuring balanced representation across tasks.

In order to overcome these obstacles, we introduce Uni-Food, a unified dataset that encapsulates various food-related information, \ie food category, ingredients, recipes, and valuable ingredient-level nutritional information for each food image. Specifically, we curate 100,000 samples from Recipe1M \cite{salvador2017recip1m} and employ ChatGPT-4 \cite{achiam2023gpt4} to generate nutritional information for each ingredient list. This information is then aggregated to derive the nutritional data for the entire dish. Furthermore, in order to ensure a high-quality gold standard set for testing, we utilize human curation to isolate a superior subset specifically designated for testing purposes.

Our proposed dataset facilitates large-scale training of LMMs for various food-related tasks within the same dataset. Nevertheless, this endeavor also presents new challenges associated with multi-task learning when fine-tuning large models. 
Mixture of Experts (MoE) \cite{huang2023lorahub, dou2023loramoe, liu2023moelora, ponti2023latent_task_skill} has been a common strategy in the field of Natural Language Processing (NLP) to handle multi-task fine-tuning. 
This method involves the utilization of multiple \textit{expert} models, each specialized in different task or segment of the data distribution. 
% The model then learns to delegate different inputs to the appropriate expert, thus facilitating effective multi-task learning.
Recently, there has been a surge in the application of MoE to LMMs.
MoE-LLaVA \cite{lin2024llavamoe}, for instance, integrates MoE into its Feed-Forward Network (FFN) layers to enhance the model's adaptability across tasks. On the other hand, LLaVA-MoLE \cite{chen2024llavamole} incorporates the MoE paradigm into the Low-Rank Adaptation (LoRA) \cite{hu2021lora} module, selecting the top-1 experts for each specific task to guarantee the spare activation of experts. However, as MoE-LLaVA incorporates experts throughout the FFN layer, the resulting increase in training parameters can be prohibitive. Conversely, although LLaVA-MoLE utilizes LoRA experts, which are more parameter-efficient, its strategy of selecting only the top expert may constrain the model's flexibility on skill leverage. This is because certain tasks might share foundational skills. For example, in the food domain, both ingredients recognition and recipe generation are sensitive to the composition of ingredients in a dish.

To address the aforementioned issues, we propose Linear Rectified Mixture of Diverse Experts (RoDE). Following LLaVA-MoLE \cite{chen2024llavamole}, we design experts from the LoRA perspective to ensure efficient use of trainable parameters. Unlike LLaVA-MoLE, which dedicates each expert to a particular task, LLaVA-RoDE conceptualizes experts as granular skill modules, allowing multiple experts to participate in a single task.
This design inspires us to develop experts with diverse capabilities—that is, we allocate varying amounts of trainable parameters to each expert in an effort to conserve GPU memory. 
% By adopting LoRA with different ranks, we are able to preserve valuable GPU memory, especially for complex tasks.
To integrate the experts, we then introduce a linear rectified router to assign tasks to the appropriate LoRA experts. The router employs a Rectified Linear Unit (ReLU) to refine its output, which confers two significant benefits: 1) it enables sparse task-skill matching, which has been proved to surpass dense activation methods in effectiveness \cite{zhou2022dense_is_better, chen2024llavamole}; 2) the simplicity of the ReLU function makes it inherently easy to optimize. 
Consequently, this leads to a sparing activation of the diverse expert set.
RoDE is engineered to optimize the allocation and usage of the model's resources, significantly bolstering the capabilities of large-scale models in multi-task learning scenarios.
Our empirical findings underscore the superior performance of RoDE in the realm of food multi-task learning challenges, validating the strength of our proposed methodology.

Our contributions can be summarized as follows:
\begin{itemize} %[leftmargin=*]
\item We introduce Uni-Food, a novel dataset encompassing a variety of food vision tasks. Building upon the substantial ingredient and recipe dataset, we further incorporate valuable ingredient-level nutritional information to facilitate relevant dietary research.

\item We propose Linear Rectified Mixture of Diverse Experts (RoDE) approach designed to effectively tackle food multi-tasks learning. RoDE approach leverages a combination of LoRA experts with varying ranks to model tasks of different complexities and employs linear rectified router to sparsely allocate these experts to appropriate tasks.

\item Experimental results demonstrate the effectiveness of our proposed approach on food multi-task learning. And ablation studies highlight the GPU memory efficiency and the sparse allocation of experts intrinsic to our RoDE model.

\end{itemize}

\section{Related Work}

\textit{Large Multimodal Models (LMMs).}
In recent years, Large Language Models (LLMs) have showcased remarkable prowess in Natural Language Processing (NLP), paving the way for breakthroughs in multimodal learning. LMMs typically integrate a pre-trained vision encoder with a LLM architecture. Subsequently, visual features undergo adaptation via a projection module, facilitating their integration into the hidden space of LLMs for joint processing with textual inputs. Through multimodal training, LMMs acquire the capability to generate responses based on both visual and textual inputs.
LLaVA \cite{liu2023llava}, for instance, introduces a vision encoder to LLMs and demonstrates significant enhancements across various vision-language tasks. Building upon LLaVA's advancements, LISA \cite{lai2023lisa} further enriches multimodal models by incorporating a segmentation module. This addition augments the model's capacity to discern fine-grained details within visual inputs, resulting in more nuanced and contextually richer representations.
GPT-4v \cite{achiam2023gpt4} stands out as one of the most powerful LMMs, capable of providing instructional data across numerous research domains. Its potency extends beyond NLP, contributing to advancements in various interdisciplinary fields.

\vspace{0.05in}
\noindent \textit{Food Multi-task Learning.}
With food playing a significant part in human life, there has been a lot of work carried out doing research in this domain. In the early stage, food classification \cite{Food101,qiu2022mining} happened to be the most popular task related, bringing up a surge in the number of relevant datasets, like Food-101~\cite{Food101}, UEC Food 256~\cite{kawano14c} and Food2K~\cite{min2023prenet}. As the tasks in the food domain developed into greater diversity, from ingredient recognition \cite{wang2022ingredient_guided_region_discovery,9794570,luo2023caclnet}, recipe generation \cite{salvador2019inversecooking, Chhikara_2024_WACV} to nutrition estimation 
 \cite{10.1145/3347448.3357172, thames2021nutrition5k}, food datasets have also grown to be more diverse and large-scale. Vireo Food-172~\cite{chen2016vireofood172} and 251 \cite{chen2020mult_task_region_ingredient} are datasets containing not only category information but also ingredient labels, Recipe1M \cite{salvador2017recip1m} and Recipe1M+~\cite{marin2019recipe1m} are datasets involving a wealth of recipe data, and Nutrition5K~\cite{thames2021nutrition5k} is a dataset specialized in high accuracy nutritional content annotation.
While all of the above methods have demonstrated their performance on food, they were all limited to address one or a few tasks, rather than integrating all tasks into a single multimodal model. To accomplish this, Yin et al. came up with FoodLMM~\cite{yin2023foodlmm}, a versatile food assistant based on LMMs that could handle a variety of tasks in the food domain. However, their method might lead to task conflict when several tasks were fine-tuned at the same time, lacking the capability to efficiently implement multi-task learning.

\begin{table}[]
\caption{\textbf{Annotation Diversity in Food Datasets.} The volume and variety of annotations differ significantly across various food datasets. Uni-Food offers a large-scale training dataset, accommodating a broad spectrum of food-related tasks.}
\label{tbl:dataset_comparison}
\resizebox{0.48\textwidth}{!}{
\begin{tabular}{lcccc}
\toprule
Dataset         & category & ingredient & recipe & nutrition \\
\midrule
Food-101        & 101k          & 0             & 0          & 0          \\
Food2K          & 1,036,564     & 0             & 0          & 0          \\
Vireo Food-251  & 169,673       & 169,673       & 0          & 0          \\
Nutrition5k     & 0             & 20k           & 0          & 20k        \\
Recipe1M        & 887,706       & 887,706       & 887,706    & 0          \\
Recipe1M+       & 13,735,679    & 13,735,679    & 887,536    & 42,713     \\
Uni-Food            & 100k          & 100k          & 100k       & 100k       \\
\bottomrule
\end{tabular}
}
\end{table}

\vspace{0.05in}
\noindent \textit{Mixture of Experts (MoE)} dynamically combines the decisions of multiple experts on the same input to improve overall performance. 
In LLMs, which typically adopt transformer architecture, MoE is often implemented within MLP layers.
% where multiple linear branches are integrated together, and a learnable router is used to manage experts
LoRAHub \cite{huang2023lorahub} initially trains a series of LoRA weights on upstream tasks. To adapt these to a downstream task, it employs a gradient-free method to search for the coefficients that will combine the pre-trained LoRA set. 
MOELoRA \cite{liu2023moelora}, on the other hand, utilizes a router conditioned on a task identifier to dynamically merge multiple LoRA outputs. In contrast, MoCLE \cite{gou2023mocle} designs a router that is conditioned on the clustering information of each individual input sample. LoRAMoE \cite{dou2023loramoe} partitions the LoRA experts into two groups and purposefully cultivates different capabilities within each group.
All these mixture-of-LoRA methods have pre-set hyperparameters that require meticulous selection, and the LoRA experts are densely combined. \cite{zhou2022dense_is_better} conducted a comparison between the dense and sparse mixtures of LoRA experts for large language models, concluding that a dense mixture yields superior performance. 
Some studies have migrated MoE designs into LMMs.
MoE-LLaVA \cite{lin2024llavamoe} introduces multiple FFNs in stead of a single one and integrates a router mechanism to sample predictions. LLaVA-MoLE \cite{chen2024llavamole} applies multiple LoRAs and activates one specifically for a task. 
However, these methods fall short in effectively handling the nuances of rational, fine-grained task-skill allocation. Our paper aims to address this gap.

\begin{figure*}[t]
\centering
\includegraphics[width=0.98\linewidth]{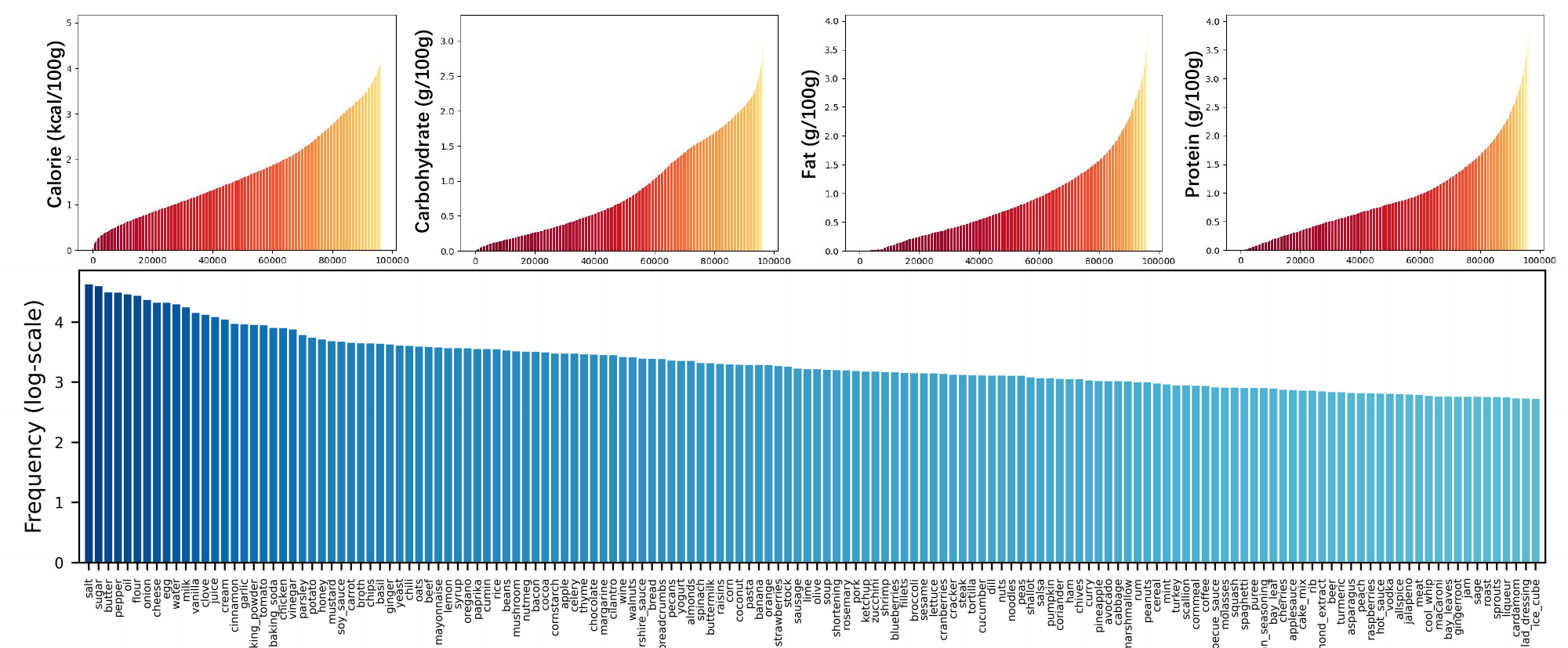}
\caption{\textbf{The nutritional statistics (per 100g) and ingredient statistics for UniFood.}
}
\label{img:sample}
\end{figure*}

\section{Dataset Construction}

In this paper, we construct a large-scale dataset called Uni-Food. Different from other publicly available food datasets~\cite{Food101, marin2019recipe1m, chen2016vireofood172, thames2021nutrition5k}, Uni-Food contains various attributes used in food-related tasks, including food category, ingredients, recipe and nutrition for each food image. To the best of our knowledge, this is the largest dataset that provides all the attributes in one dataset. 
Table \ref{tbl:dataset_comparison} summarizes the tasks and sample sizes of existing primary datasets and our Uni-Food dataset.

\subsection{Attribution}
\begin{figure}[t]
\centering
\includegraphics[width=0.8\linewidth]{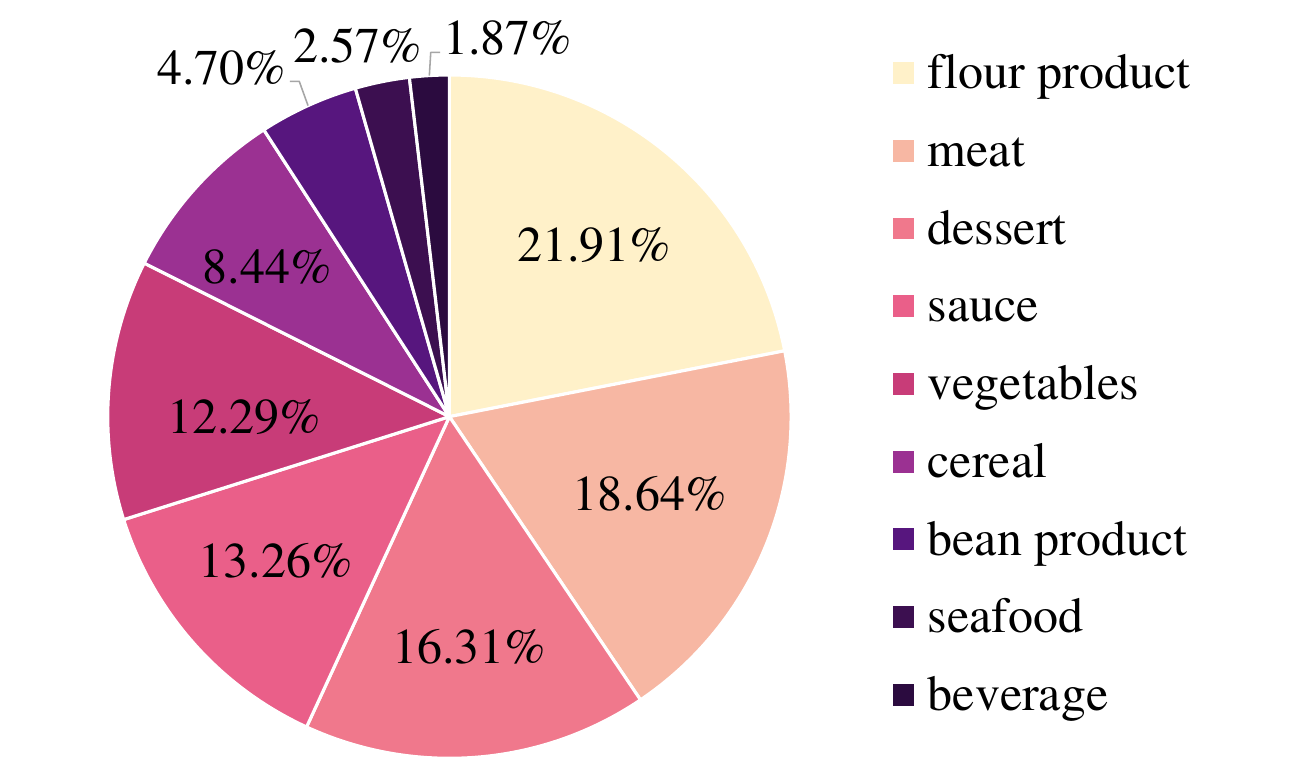}
\caption{\textbf{UniFood Dataset Statistics.} The category distribution of the dataset.
}
\label{img:food_distri}
\end{figure}

Our objective is to construct a unified and comprehensive dataset containing rich information relevant to food, encompassing the following key attribution for each image.
\textbf{Category:} Classifying each food item into specific categories to facilitate organization and categorization within the dataset.
\textbf{Ingredients Information:} Providing a thorough breakdown of the ingredients used in each dish, including their names and quantities.
\textbf{Cooking Instructions:} Offering step-by-step instructions on how to prepare each dish, ensuring clarity and completeness for easy reproduction.
\textbf{Nutrition Information:} Incorporating detailed nutritional data for each dish, such as macronutrient content (e.g., carbohydrates, proteins, fats), micronutrients, and total calories.
An intuitive sample demonstration of these attributions is shown in Figure \ref{img:sample}. The distribution across various categories is visualized in Figure \ref{img:food_distri}

% \subsection{Dataset collection}
% Given the vastness of Recipe1M, comprising extensive recipe and image pairs, we have undertaken a random selection of 100,000 samples from this dataset. Each selected sample includes an image, a title, an ingredient list, and cooking instructions. The distribution across various categories is visualized in Figure \ref{img:food_distri}.

\begin{figure*}[t]
  \centering
  \includegraphics[width=1\linewidth]{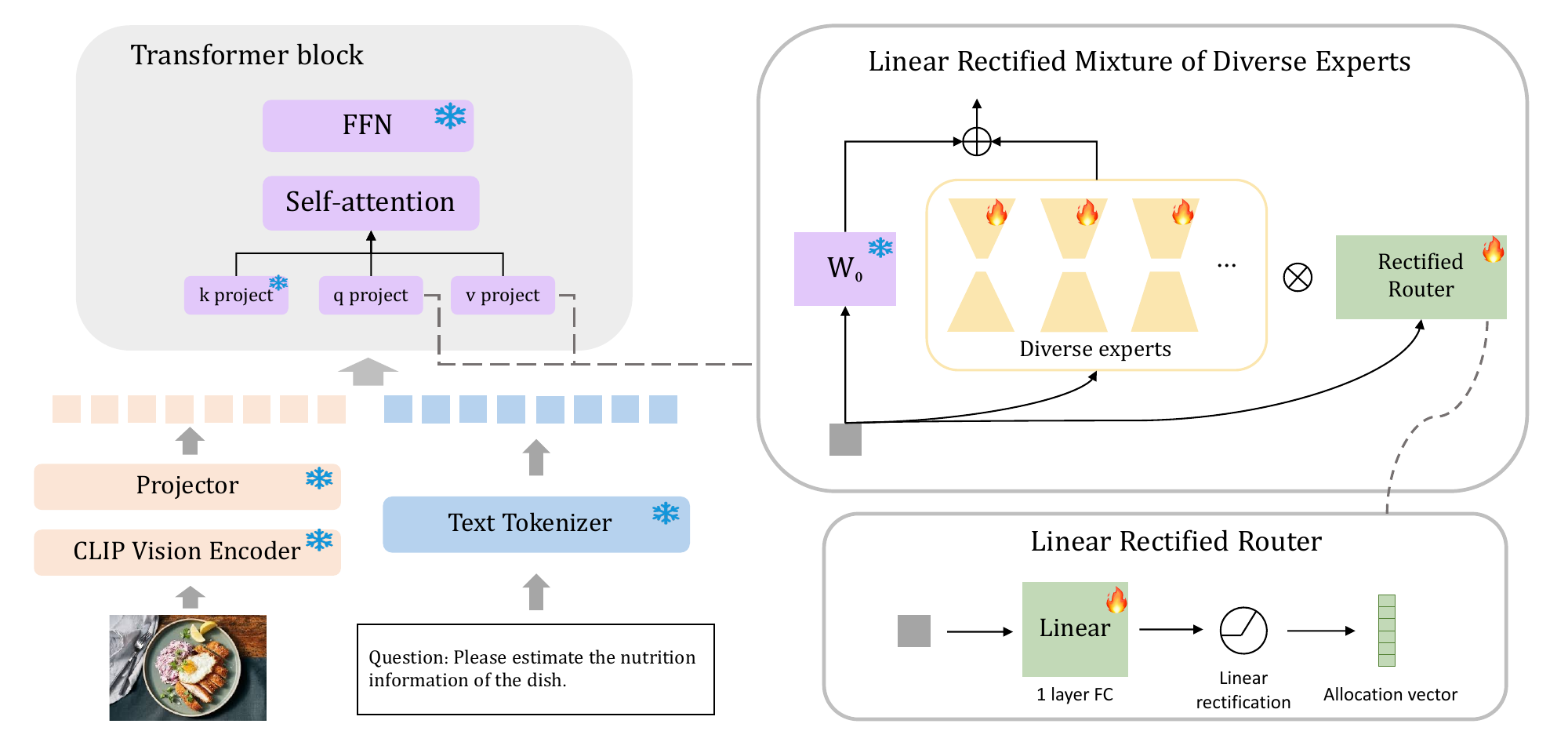}
  \caption{The illustration of our proposed Linear Rectified Mixture of Diverse Experts (RoDE) approach. We use LLaVA \cite{liu2023llava} as the foundational LMM. RoDE module is incorporated into both the query projection layer and the value projection layer of each transformer block.}
  \label{img:rode}
\end{figure*}

\subsection{Nutrition Labeling}
As the ingredient and recipe information can be easily collected from Recipe1M+ \cite{marin2019recipe1m}, we proceed to annotate the nutrition content.
To acquire precise nutritional information, we feed both the food image and the ingredients with their respective amounts into ChatGPT4-vision \footnote{We intentionally omit recipe information to enhance processing efficiency, as we have observed that including recipes increases query time without considerably improving precision.}. The output text is meticulously processed to extract the nutritional values. 
Leveraging the capabilities of ChatGPT4-vision, we collect detailed nutrition information at the ingredient level, encompassing metrics such as mass, fat content, protein content, carbohydrate content, and energy content. Importantly, the ingredient list in recipes contains the quantity for each ingredient, along with the food image,  ChatGPT4-vision is capable of generating high-quality ingredient-level nutrition information according to our manual check based on USDA~\footnote{https://fdc.nal.usda.gov/}. For detailed prompt design, please refer to the Sec. A of the supplementary materials. Subsequently, the ingredient-level nutrition information is then aggregated to derive the comprehensive nutrition profile of the entire dish. Through this process, we obtain a holistic understanding of the nutritional composition of the dish, enabling precise estimation and analysis of its nutritional value.

\subsection{Golden set selection}
To ensure accurate evaluation, we construct a precise gold standard as the test set. To accomplish this goal, we manually filter out inaccurate samples by cross-referencing their nutritional information with the USDA database. Additionally, we eliminate certain overrepresented categories to achieve a balanced distribution of sample categories.
% This rigorous selection process ensures that our testing dataset comprises high-quality samples that closely align with real-world nutritional data, enhancing the reliability and validity of our evaluation outcomes.
% we employ Recip1M+ as a reference to select valuable and precise samples. Specifically, we conduct a comparative analysis of the nutrition information between Uni-Food and Recip1M+\cite{marin2019recipe1m}. Samples with differences in nutrition content below a predefined threshold, \ie 20\%, are retained. 
% Then, we manually filter inaccurate samples by comparing the nutrition information from the USDA database and balance the categories of the selected samples.
% This rigorous selection process ensures that our testing dataset comprises high-quality samples that closely align with real-world nutritional data, enhancing the reliability and validity of our evaluation outcomes.

\subsection{Nutrition annotation quality}
We ensured high-quality nutrition annotation in the ChatGPT-4V by supplying it with both the quantity and name of each ingredient from the recipe, as verified through our earlier manual checks. To evaluate the disparity between ChatGPT annotation and human annotation, we randomly select 1,000 entries from our golden nutrition test set and prompt ChatGPT-4V to generate their nutrition information. The results, detailed in Table \ref{tbl:qua_ana}, show minimal discrepancies between the human-verified values and those generated by ChatGPT-4V, demonstrating the high quality of our annotations. For further insights, case studies on nutritional annotation are discussed in Sec. B of the supplementary material.

% The manual verification of nutrition annotations in UniFood is quite costly. 
\begin{table}[h!]
    \centering
    \caption{\small{Gap between Human and ChatGPT-4V Annotations.}}
    \label{tbl:qua_ana}
    \resizebox{0.4\textwidth}{!}{%
    \begin{tabular}{lcccc}
        \toprule
        Metric & Mass (g) & Calories (kcal) & Fat (g) & Protein Value (g) \\
        \midrule
        MAE  & 13.22 & 54.42 & 2.67 & 2.39   \\
        pMAE & 1.77\% & 2.7\% & 2.59\% & 6.91\%   \\
        \bottomrule
    \end{tabular}
    }
\end{table}

\section{Method}
% In this section, we initially provide a preliminary overview of the problem formulation for multi-task tuning of LMMs, along with commonly used techniques, \ie Low-Rank Adaptation (LoRA) \cite{hu2021lora} and Mixture of Experts (MoE) \cite{dou2023loramoe} for addressing this challenge. Subsequently, we introduce our Non-linear Allocated Diverse Experts approach.
In this section, we first provide a preliminary overview of the problem formulation for multi-task tuning of LMMs, as well as the commonly used techniques, \ie Low-Rank Adaptation (LoRA) \cite{hu2021lora} and Mixture of Experts (MoE) \cite{dou2023loramoe} for addressing this problem.
Subsequently, we introduce our Linear Rectified Mixture of Diverse Experts approach in detail.

\subsection{Preliminary}
\subsubsection{Problem Formulation.}
A Large Multi-modal Model (LMM) can be constituted by a vision encoder and a Large Language Model (LLM), as shown in Figure \ref{img:rode}. Normally, LMM is initially pre-trained with tremendous amounts of data, and then fine-tuned to adapt to downstream tasks. 
The utilization of multi-modal documents for Supervised Fine-Tuning (SFT) \cite{radford2018bert} is a common practice in the fine-tuning of LMMs.
% Supervised Fine-Tuning (SFT) \cite{radford2018bert} is commonly used to fine-tune LMM, leveraging multi-modal documents for auto-regressive training.

Let us denote the set of multimodal documents as \(D\), where \(D = \{(I_i, T_i)\}_{i=1}^{M}\), where \(I_i\) signifies the image, and \(T_i\) represents the associated set of tasks with that image. \(M\) stands for the total number of documents.
Each task set \(T_i\) encompasses sequence-specific tasks \( \left \{ (q_i^j, a_i^j) \right \} _{j=1}^{\mathcal{T}}\), where \(q_i^j\), \(a_i^j\) is the question and answer for task \(j\), and \(\mathcal{T}\) is the number of task types.
The primary objective of SFT is to using \(D\) to fine-tune the LMM so that it can provide corresponding answers to given questions based on an image.
More specifically, for an image \(I_i\) and a question \(q_i^j\), 
the training objective is to maximize the probability \(p(a_i^j|I_i, q_i^j, \theta)\), where \( \theta \) represents the trainable parameters of the large model.

% the training objective is to minimize the discrepancy between the prediction probability \(p(a_i^j|I_i, q_i^j, \theta)\) and the one-hot encoded ground truth answer tokens, where \( \theta \) is trainable parameters.

\vspace{0.1in}
\subsubsection{Low-Rank Adaptation (LoRA).}
Fully tuning large models can be resource-intensive. Parameter-Efficient Fine-Tuning (PEFT) \cite{ding2023peft} introduces additional adapters to effectively customize large models for downstream tasks with minimal resource overhead. 
% Our study primarily employs Low-Rank Adaptation (LoRA) \cite{hu2021lora} as the adapter.
Let $W$ represent the weight of a linear layer $L$ in the large-scale model, which is initially set as $W_0$ after pre-training. In PEFT, $W_0$ remains frozen in order to preserve the acquired world knowledge. To facilitate adaptation to downstream tasks, a learnable branch linear adapter, denoted as $\Delta W$, is introduced to modify the initial frozen linear weights $W_0$. Consequently, the output of the adapted linear layer $L$ can be expressed as $W = W_0 + \Delta W$.
LoRA \cite{hu2021lora} further decomposes $\Delta W$ into two matrices, $A$ and $B$, where the connection rank is considerably smaller than that of $W_0$. LoRA allows the model to adapt to downstream tasks while reducing the number of training parameters.

\subsubsection{Mixture of Experts (MoE).} \label{sec:lora_moe}
The Mixture of Experts (MoE)~\cite{dou2023loramoe} paradigm of LoRA is proposed to adapt large models to multiple downstream tasks by employing multiple LoRAs in the MLP layers. Each LoRA module can be considered as an \textit{expert}, wherein all experts receive the same input and combine their outputs to improve overall performance.
Let $R = \left \{ A_i, B_i\right \} _{i=0}^{N} $ denote a series of LoRA modules, where $N$ signifies the number of LoRAs. We use $h(\cdot)$ to denote the linear router, which takes the dimension of the input feature $x$ and outputs the allocation vector of $R$. 
The output of layer $L$ incorporating the MoE module can be represented as:
\begin{eqnarray}
x' & = & W_0 x + \sum_{i=0}^{N}  \frac{\alpha}{r} \sigma(h_i)B_i^rA_i^rx. 
\end{eqnarray}

\noindent Here, $\sigma$ stands for the softmax operation, $\alpha$ is a hyperparameter, and $x' $ is the output feature.

\subsection{Linear Rectified Mixture of Diverse Experts}
Our proposed RoDE framework incorporates a variety of experts, each with distinct capabilities, and a linear rectification router to integrate the contributions of these experts. The overall structure of the framework is depicted in Figure \ref{img:rode}. A comprehensive illustration of the framework is provided in the subsequent sections.

% \vspace{0.05in}
\subsubsection{Diverse Capability Experts}

Typically, within the Mixture of Experts (MoE) framework, all experts share the same architecture. In the case of LoRA, for example, each expert is assigned the same rank. However, this uniformity assumes equal capabilities across all experts. This implies that for more complex tasks, each expert might need to possess a large number of parameters to adapt effectively. However, this could be prohibitively demanding in terms of GPU memory requirements.

To mitigate the issue of GPU memory constraints, we conceptualize the experts as fine-grained skill modules. The key idea is that a task may activate a combination of these modules, and these modules can be shared across various tasks. This modular design intuitively leads us to develop LoRA experts with different capabilities tailored to tasks of varying complexity. Drawing inspiration from Low-Rank Adaptation \cite{hu2021lora}, which suggests that a low-rank adapter may be sufficient for certain tasks, we create LoRAs with varying ranks. Consequently, the resulting skill space comprises both high-rank and low-rank LoRA experts, providing greater flexibility and efficiency in addressing a diverse set of tasks.

To construct such skill space, we configure a heterogeneous set of experts, represented as $R = \left \{ A_i^{r_i}, B_i^{r_i}   \right \}_{i=0}^{N}$, where the rank $r_i$ of $i-th$ LoRA module may vary from the others. Therefore, we can establish a skill space composed of various LoRAs, each tailored to accommodate tasks of different complexity levels.

\begin{figure}[t]
\centering
\includegraphics[width=\linewidth]{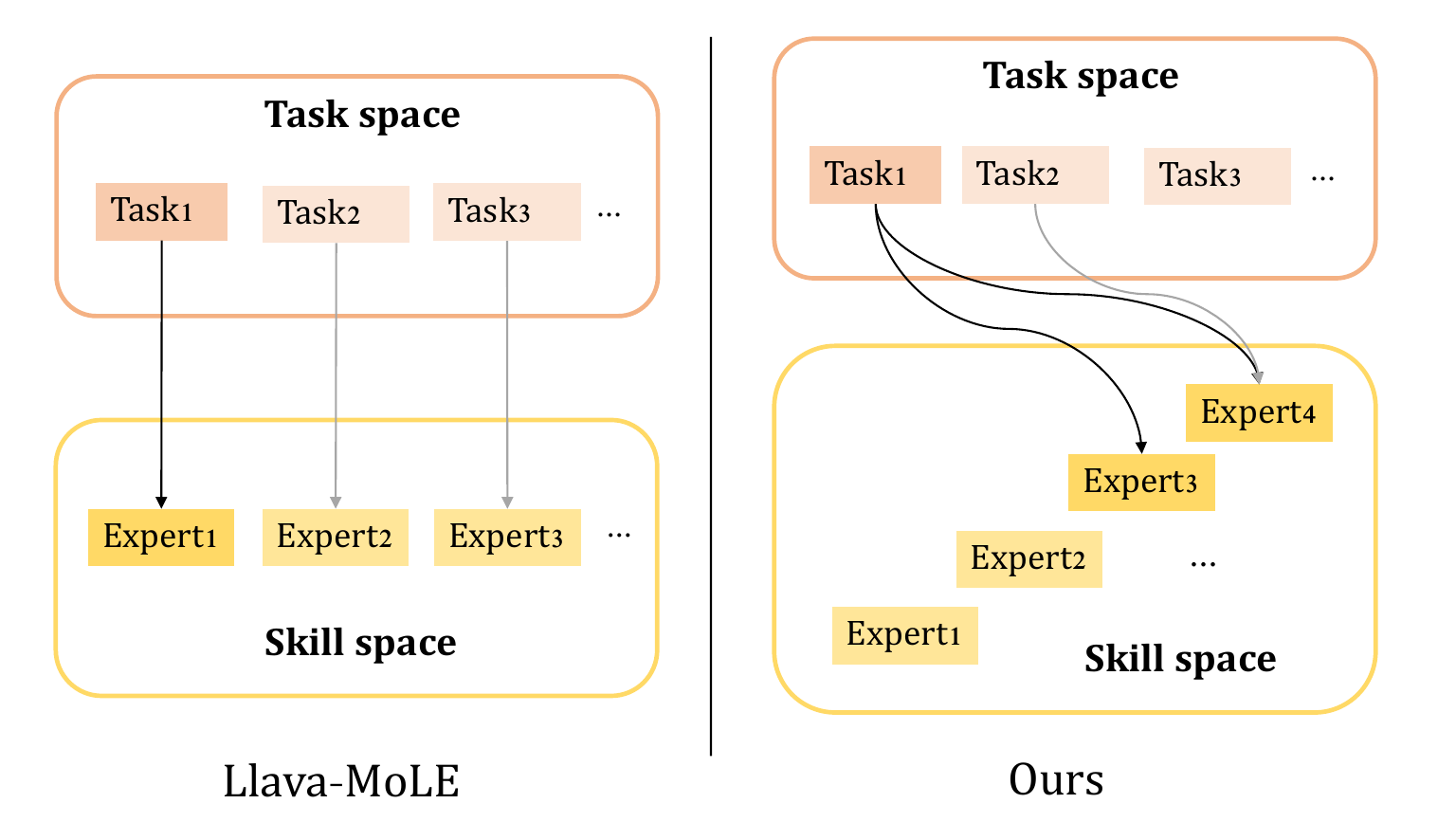}
\caption{\textbf{Comparison between LLaVA-MoLE \cite{chen2024llavamole} and our proposed RoDE.} In our RoDE framework, experts are utilized across all tasks. For individual tasks, expert activations are optimized to achieve sparsity.
}
\label{img:task-skill_space}
\end{figure}

% \vspace{0.05in}
\subsubsection{Linear Rectified Router}
The router consolidates the contributions of each expert based on the demands of the specific task. Previous research \cite{zhou2022dense_is_better} has shown that the use of sparse mixtures of LoRA experts outperforms dense experts in the context of large language models. LLaVA-MoLE \cite{chen2024llavamole} presents a top-1 selection strategy, which ensures the sparsity of expert selection in LMM.
While some methodologies in the Natural Language Processing (NLP) field adopt a “soft” approach to combine expert outputs—for instance, \cite{dou2023loramoe} utilizes softmax and \cite{ponti2023latent_task_skill} employs Gumbel softmax—these techniques are not as sparse and can be challenging to optimize.

In contrast, our methodology adopts a novel approach by utilizing the Rectified Linear Unit (ReLU) \cite{nair2010RELU} to rectify the output of the routers, thereby encouraging the sparse learning of LoRA expert activations. A visual illustration of our routing strategy is shown in Figure \ref{img:task-skill_space}. Leveraging the intrinsic properties of the Rectified Linear Unit (ReLU), our approach benefits from a simplified optimization landscape and fosters sparsity within the network, which enables our model to boost both efficiency and efficacy.

Let $\gamma$ represent the linear rectification operation, $\gamma(x) = \max (x, 0)$. 
We utilize ReLU to rectify the linear router, resulting in an adjusted linear output that can be expressed as follows:
\begin{eqnarray}
x'  = W_0x + \sum_{i=0}^{N} \frac{\alpha}{r_i}\gamma(h_i) B_i^{r_i}A_i^{r_i}  x ,
\end{eqnarray}
where $(A_i^{r_i}, B_i^{r_i})$ represents the operation performed by the $i$-th LoRA expert. The summed output from the RoDE module is then added to the frozen linear output and forwarded to the next module.

This linear rectification mechanism, enhances the sparsity of skill selection, facilitating efficient utilization of expertise across a broad spectrum of tasks. Overall, diverse LoRA experts and linear rectified arrangement yield a fine-grained task-skill arrangement space, contributing to the optimized allocation and utilization of resources within the model.

\begin{table*}[t]
\centering
\caption{\textbf{Comparative results for ingredient recognition, recipe generation and nutrition estimation on the Uni-Food Dataset.} The FoodLMM+MoE is derived by integrating a Mixture of Experts (MoE) paradigm into the FoodLMM model.}
\label{tbl:uni-food}
\resizebox{\textwidth}{!}{
\begin{tabular}{lcccccccccc}
\toprule
\multirow{2}{*}{Method}  & \multicolumn{2}{c}{\textbf{Ingredient recognition}} & \multicolumn{2}{c}{\textbf{Recipe generation}} & \multicolumn{6}{c}{\textbf{Nutrition estimation (pMAE)}}     \\
& IoU $(\uparrow )$  &F1 $(\uparrow )$              & SacreBLEU $(\uparrow )$          & Rouge-L $(\uparrow )$          & Mass $(\downarrow )$& Energy $(\downarrow )$& Fat $(\downarrow )$& Proteins $(\downarrow )$& Carb $(\downarrow )$& Avg.$(\downarrow )$ \\
\midrule
QwenVL \cite{bai2023qwenvl} &20.01 & 30.51 & 7.97 & 39.19   & 49.99 & 48.12 & 65.19 & 62.95 & 49.41 & 55.13 \\
FoodLMM \cite{yin2023foodlmm}    &   22.59       & 32.98           &     7.59           &     36.44              &  48.49    &  47.72      & 61.49    &   \textbf{56.61}      &  51.25   &  53.11    \\
FoodLMM+MoE   &    24.54  & 35.21   &     8.7     &    38.5       &   49.23   &  48.24    &  \textbf{60.59}   &   58.37      &  51.87   &  53.66   \\
RoDE (ours)    &   \textbf{26.86}     & \textbf{37.67}        &       \textbf{9.66}     & \textbf{40.11}         &      \textbf{48.18}          &      \textbf{46.96}             &  60.96    &   58.16     &  \textbf{48.65}          &   \textbf{52.58}     \\
\bottomrule
\end{tabular}
}
\end{table*}

\begin{table*}[t]
\centering
\caption{\textbf{Results on Food-101, VireoFood-172 and Nutrition 5k.} We utilize Food-101 for the classification task, VireoFood-172 for the ingredient recognition task, and Nutrition 5k for nutrition estimation task. The term “FoodLMM-s1" refers to the version of FoodLMM that has been fine-tuned on the target dataset. Similarly, “RoDE-ft" denotes the fine-tuned version of RoDE.} 
\label{tbl:ft}
% \resizebox{\textwidth}{!}{
\begin{tabular}{lcccccccccc}
\toprule
\multirow{2}{*}{Method}  & \textbf{Food 101} & \multicolumn{2}{c}{\textbf{VireoFood-172}} & \multicolumn{6}{c}{\textbf{Nutrition 5k}}     \\
& Acc $(\uparrow )$                 & IoU $(\uparrow )$          & F1 $(\uparrow )$          & Mass $(\downarrow )$& Energy $(\downarrow )$& Fat $(\downarrow )$& Proteins $(\downarrow )$& Carb $(\downarrow )$& Avg.$(\downarrow )$ \\
\midrule
PRENet \cite{min2023prenet} & 91.13 & - & - & - & - & - & - & - & - \\
CACLNet \cite{luo2023caclnet} & - & - & 65.71  & - & - & - & - & - & - \\
FoodLMM-s1 \cite{yin2023foodlmm} & 93.93 & 56.94 & 68.97 & 38.96 & 54.04 & 69.84 & 54.88    & 51.3  & 53.8  \\
FoodLMM-s1+MoE & \textbf{94.1}  & 61.38  & 69.36  & 39.31 & 53.6  & 67.14 & \textbf{52.99}    & 49.59 & 52.53 \\
RoDE-ft (ours) & 94.02 & \textbf{65.95} & \textbf{73.54}  & \textbf{38.45} & \textbf{52.47} & \textbf{67.1}  & 53.92    & \textbf{47.84} & \textbf{51.96}  \\

\bottomrule
\end{tabular}
% }
\end{table*}

\subsection{Optimization Objects and Inference}
Following previous LMM-based methods \cite{liu2023llava}, the input images are transformed into image tokens and concatenated with text tokens to send to LLM. The training process adheres to the autoregressive model, predicting the next token based on the input tokens. We employ the standard cross-entropy loss as the optimization objective to train our model.

During inference, for each task, we input the image along with the corresponding question into the model. The model's output tokens are then converted into words to formulate the answer.
% For tasks requiring numerical metric calculations, \ie nutrition estimation, we employ regular expressions to extract values for specific nutritional elements.

% \begin{table*}[]
% \resizebox{\textwidth}{!}{
% \begin{tabular}{lcccccccccc}
% \toprule
% \multirow{2}{*}{Method} & \textbf{Food 101} & \multicolumn{2}{c}{\textbf{VireoFood 172}} & \multicolumn{6}{c}{\textbf{Nutrition 5k}} \\
%  & Acc & IoU & F1 & Mass & Energy & Fat & Carb & Proteins & Avg. \\
%  \midrule
% PRENet \cite{min2023prenet} & 91.13 & - & - & - & - & - & -  & - & - \\
% CACLNet \cite{luo2023caclnet} & - & - & 65.71 & - & - & - & - & - & - \\
% 2D Direct Prediction \cite{thames2021nutrition5k} & - & - & - & 70.6 / 26.1\%  & 40.4 / 18.8\% & 5.0 / 34.2\% & 6.1 / 31.9\% & 5.5 / 29.5\% & 25.52 / 28.1\% \\

% FoodLMM-s1 \cite{yin2023foodlmm} & 93.93 & 56.94 & 68.97 &  67.3 / 26.6\% & 38.8 / 20.2 \% & 5.5 / 40.4 \% & 6.1 / 31.9 \% & 4.2 / 26.2 \% & 24.38 / 29.1\%  \\
% RoDE (ours) &  &  &  &  &  &  &  &  \\
% \bottomrule
% \end{tabular}
% }
% \caption{\textbf{Results on common datasets.} We utilize Food101 for the classification task, VireoFood 172 for the ingredient recognition task, and Nutrition 5k for the nutrition estimation task.}
% \label{tbl:other_dataset}
% \end{table*}

\section{Experiment}
In this section, we first present our experimental setup. Then we elaborate experimental results of multiple food tasks based on multi-task learning.
% conducted on multi-task learning in the context of food-related tasks.
Following this, we perform an ablation study to evaluate the impact and effectiveness of our main components.

% \begin{table}[]
% \centering
% \caption{\textbf{Results on Food-101 and VireoFood-172.} We utilize Food-101 for the classification task, VireoFood-172 for the ingredient recognition task.}
% \label{tbl:other_dataset}
% % \resizebox{\textwidth}{!}{
% \begin{tabularx}{0.45\textwidth}{lCCC}
% % \begin{tabular}{lcccc}
% \toprule
% \multirow{2}{*}{Method} & \textbf{Food-101} & \multicolumn{2}{c}{\textbf{VireoFood-172}}  \\
%  & Acc & IoU & F1  \\
%  \midrule
% PRENet \cite{min2023prenet} & 91.13 & - & -  \\
% CACLNet \cite{luo2023caclnet} & - & - & 65.71  \\

% FoodLMM-s1 \cite{yin2023foodlmm} & 93.93 & 56.94 & 68.97 \\
% RoDE (ours) & \textbf{94.02} & \textbf{65.95} & \textbf{73.54}  \\
% \bottomrule
% % \end{tabular}
% \end{tabularx}
% % }
% \end{table}

% \begin{table}[]
% \caption{\textbf{Nutrition estimation results on Nutrtion5k dataset.} The evaluation metric employed is the percentage Mean Absolute Error (pMAE).}
% \label{tbl:nutrition5k}
% \resizebox{0.5\textwidth}{!}{
% \begin{tabular}{lcccccc}
% \toprule
% Nutrition5k            & Mass $(\downarrow )$ & Cal $(\downarrow )$  & Fat $(\downarrow )$  & Proteins $(\downarrow )$ & Carb $(\downarrow )$  & Avg. $(\downarrow )$ \\
% \midrule
% FoodLMM\cite{yin2023foodlmm}     & 38.96 & 54.04 & 69.84 & 54.88    & 51.3  & 53.8  \\
% FoodLMM+MoE & 39.31 & 53.6  & 67.14 & \textbf{52.99}    & 49.59 & 52.53 \\
% RoDE (ours) & \textbf{38.45} & \textbf{52.47} & \textbf{67.1}  & 53.92    & \textbf{47.84} & \textbf{51.96}  \\
% \bottomrule
% \end{tabular}
% }
% \end{table}

\subsection{Experiment Setup} %- Training stage and dataset
Our approach commenced by utilizing the pre-trained weights of the LLaVA-Lightning-7B-v1-1 \cite{liu2023llava} model, subsequently applying SFT on the Uni-Food and Nutrition5k datasets \cite{thames2021nutrition5k}. 
In our RoDE model, we employ a heterogeneous array of experts with LoRA ranks of [2, 4, 8, 16].
Our primary baseline is \textbf{FoodLMM} \cite{yin2023foodlmm}, which to our knowledge is the only versatile LMM specialized in food-related tasks.
Additionally, we compare with \textbf{FoodLMM+MoE}, which substitutes the plain LoRA module with a Mixture of Experts (MoE) module. Each MoE module consists of four LoRAs with a rank of 8. The routing mechanism for the MoE module is a linear layer followed by a softmax operation, as detailed in Section \ref{sec:lora_moe}.
Furthermore, we explore the capabilities of \textbf{QwenVL} \cite{bai2023qwenvl}, a LMM that is not ideally suited for the food domain. We fine-tune QwenVL on the UniFood dataset using pretrained weights.
Importantly, for simplicity, we omit the specialized token tags and additional heads used by FoodLMM specifically for nutrition estimation, and instead directly convert the output tokens from the LMM into text to generate nutritional predictions.
For the standardization of the ingredient vocabulary, we adopt the same clustering protocol used for ingredients in InverseCooking \cite{salvador2019inversecooking}.

\subsubsection{Tasks and Evaluation Metrics}
% \noindent \textbf{Tasks and Evaluation Metrics.}
The tasks and corresponding metrics in our experiments are as follows.
% \textbf{Food Classification:} This task entails predicting the title or name of the food depicted in an image. We evaluate performance using classification accuracy (Acc) as the metric.
\textbf{Ingredient Recognition:} The objective here is to identify and list the ingredients present in a dish shown in an image. We assess performance using Intersection over Union (IoU) as the metric.
\textbf{Recipe Generation:} The goal of this task is to generate cooking instructions for the dish captured in an image. We measure performance using the commonly used text generation metrics in NLP, SacreBLEU \cite{bleu} and Rouge-L \cite{Rouge}.
\textbf{Nutrition Estimation:}
This task involves estimating a range of nutritional parameters from a dish's image, including the total food mass, total energy, total fat, total carbohydrates, and total protein content. Energy is measured in kilocalories, while the remaining parameters are quantified in grams. 
We employ the mean absolute error as a percent of the respective mean for that field (pMAE) \cite{thames2021nutrition5k} for the metric of nutrition estimation.
% In line with \cite{thames2021nutrition5k}, we employ the Mean Absolute Error (MAE) and mean absolute error as a percent of the respective mean for that field (pMAE).

\begin{figure}[t] 
\centering
\includegraphics[width=0.9\linewidth]{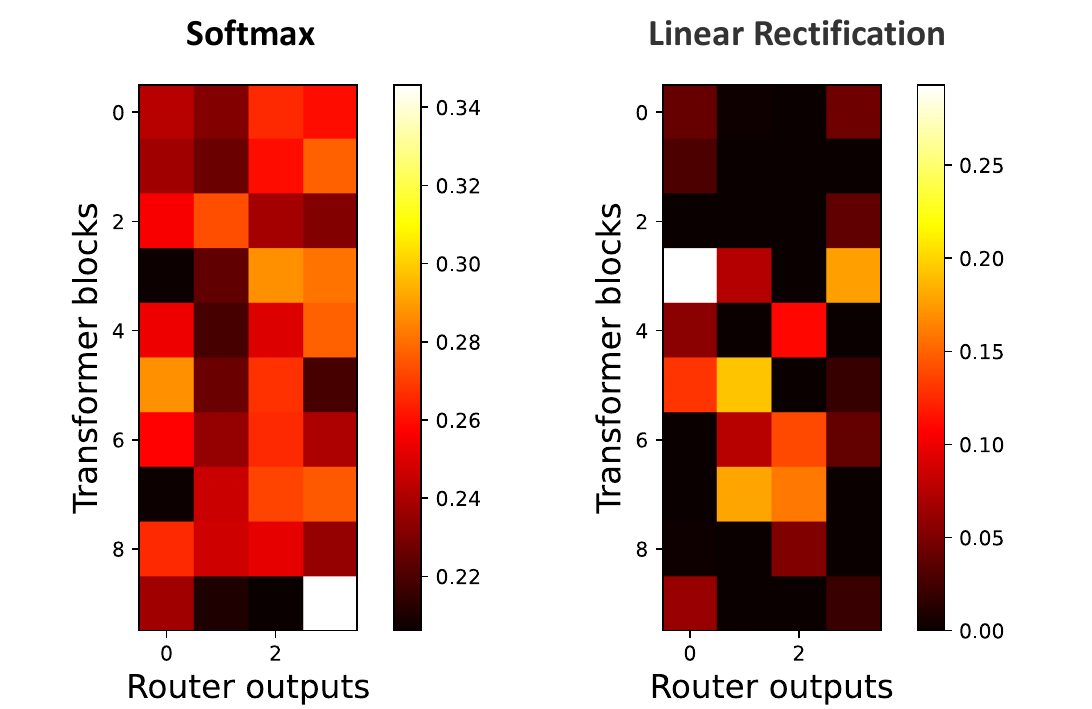}
\caption{\textbf{Comparison of Sparsity Heatmaps for Expert Activation.} This heatmap visualizes the disparity in task allocation sparsity between routers using the softmax strategy and those employing our proposed linear rectification method. The y-axis denotes the middle transformer blocks, while the x-axis represents the output of the routers.
} 
\label{img:sparsity}
\end{figure}

\begin{table*}[]
\centering
\caption{\textbf{Ablation Study on MoE.} The integration of MoE can enhance the LMM, not solely due to the introduction of additional LoRA parameters} 
\label{tbl:abla-moe}
% \resizebox{\textwidth}{!}{%
\begin{tabular}{l|c|c|c|cccc}
\toprule
Settings   & LoRA rank     & Routing strategy & MoE & SacreBLEU ($\uparrow$ ) & Rouge-L ($\uparrow$ ) & Ingredient IoU ($\uparrow$ ) & pMAE ($\downarrow$ ) \\
\midrule
Var1 & 8     &  -  & $\times$ & 7.59    &  36.44   &  22.59     &   \textbf{53.11}        \\
Var2 & 32 & - & $\times$  & 8.39 & 36.68 & 22.7 & 54.14 \\
Var3 & 8, 8, 8, 8 & Top-1  & \checkmark  & 8.28   &  37.2   &  24.08    &    53.53      \\
Var4  & 8, 8, 8, 8 &   Softmax  & \checkmark  & \textbf{8.7}   &  \textbf{38.5}   &  \textbf{24.54}    & 53.66 \\
\bottomrule
\end{tabular}
% }
\end{table*}

\begin{table*}[]
\centering
\caption{\textbf{Comparison of Three Routing Strategies.} Our proposed Linear Rectified Router demonstrates superior performance on both homogeneous and heterogeneous combinations of experts.}
\label{tbl:abla-router}
% \resizebox{\textwidth}{!}{%
\begin{tabular}{l|c|c|cccc}
\toprule
Settings   & LoRA rank    & Routing strategy  & SacreBLEU ($\uparrow$ ) & Rouge-L ($\uparrow$ ) & Ingredient IoU ($\uparrow$ ) & pMAE ($\downarrow$ ) \\
\midrule
Var1 & 8, 8, 8, 8 & Top-1  & 8.28   &  37.2   &  24.08    &    53.53      \\
Var2  & 8, 8, 8, 8 &   Softmax   & 8.7   &  38.5   &  24.54    &   53.66       \\
Var3  & 8, 8, 8, 8 &   LR  &  9.48   &  40.03   & 26.7  &   52.75       \\
\midrule
Var4  & 2, 4, 6, 8 &  Softmax   & 8.66  & 38.15   &  24.54   &  53.22       \\
Var5  & 2, 4, 6, 8 &  LR   & 9.4   &  39.67   &   26.76    &    52.63       \\
\bottomrule
\end{tabular}
% }
\end{table*}

\subsubsection{Implementation Details}
% \noindent \textbf{Implementation Details.} %- model architecture and hyperparameter
In this study, we utilize LLaVA-7B \cite{liu2023llava} as the base LMM, wherein a CLIP ViT-L model serves as the vision encoder, and Llama 7B \cite{touvron2023llama2} serves as the Language Model. For the CLIP ViT-L encoder, the input image resolution is set at 336x336, which is then divided into 14 patches. Subsequently, a two-layer MLP is used to transform these image patches into 576 tokens.
For efficiency, the LoRA module is added to the query and key projection layer. 
Unless specifically stated, the rank of LoRA is set to 8. 
Hyperparameter $\alpha$ is set to 16, and the LoRA dropout rate is set to 0.05.

During the training process of all our experiments, the weights of vision encoder and Llama are kept constant, with only the added LoRA modules being trainable.
We employ the AdamW optimizer \cite{loshchilov2017adamw} and use the WarmupDecayLR learning rate scheduler. The initial learning rate is set to 0.0003, weight decay is set to 0, and we perform 100 warm-up iterations.
In the training stage, we use a batch size of 4, and gradient updates are computed every 10 batches. The training process is distributed across 4 RTX4090 GPUs. The training epoch is set to 1.

% \subsection{Result on Uni-Food}
\subsection{Performance Comparison}

\subsubsection{Results on UniFood}
We conduct a comparative analysis among our model, QwenVL, and two versions of FoodLMM: a base version equipped with plane LoRA, and an enhanced version, FoodLMM+MoE, equipped with four 8-rank LoRA and a linear router with softmax. These methods are evaluated across three tasks: ingredient recognition, recipe generation, and nutrition estimation. The summarized outcomes in Table \ref{tbl:uni-food} reveal that FoodLMM outperforms QwenVL in ingredient recognition and nutrition estimation tasks but falls short in recipe generation. The integration of the Mixture of Experts (MoE) framework significantly bolsters FoodLMM's capabilities. Specifically, the FoodLMM+MoE version, which employs the classical MoE approach, achieves an 8.6\% enhancement in Intersection over Union (IoU) for ingredient recognition. For recipe generation, it exhibits a notable 14.6\% increase in SacreBLEU and a 5.6\% improvement in Rouge-L scores. Nonetheless, there is a marginal decrease of 1\% in the nutritional estimation performance. Overall, these comparative results underscore the benefits of incorporating the MoE paradigm into FoodLMM.

Furthermore, our model, RoDE, significantly outperforms the traditional MoE design, achieving the highest scores across all evaluated metrics. In the ingredient recognition task, RoDE surpasses FoodLMM+MoE by a notable margin of 9.5\%. For recipe generation, RoDE shows an 11\% higher SacreBLEU score and a 4.2\% improvement in Rouge-L score compared to FoodLMM+MoE. In the area of nutrition estimation, RoDE nearly tops the charts for all nutrient elements and secures the leading position in terms of average performance.
These experimental results conclusively affirm the superiority of our proposed RoDE approach in food multi-task learning.

% \subsection{Result on Nutrition5k}
% Additionally, we conduct further experiments on Nutrition5k ~\cite{thames2021nutrition5k} dataset, adhering to the methodology established by FoodLMM, where only image information was utilized for training (without depth information). The outcomes, as presented in Table \ref{tbl:nutrition5k}, unequivocally demonstrate that our methods consistently surpass both FoodLMM and traditional MoE variations in performance.

\subsubsection{Results on other datasets}
Additionally, we extend our evaluation to specific single tasks, including food classification, ingredient recognition, and nutrition estimation, utilizing distinct datasets for each task: the \textbf{Food-101} dataset \cite{Food101} for food classification, the \textbf{VireoFood-172} dataset \cite{chen2016vireofood172} for ingredient recognition, and the \textbf{Nutrition 5k} dataset \cite{thames2021nutrition5k} for nutrition estimation.
For the Food-101 dataset, our method is benchmarked against state-of-the-art methods including \textbf{PRENet} \cite{min2023prenet} and \textbf{FoodLMM} \cite{yin2023foodlmm}. In the case of the VireoFood-172 dataset, we draw comparisons with \textbf{CACLNet} \cite{luo2023caclnet} and \textbf{FoodLMM}.
Regarding the Nutrition 5k dataset, we adhere to the methodology established by FoodLMM, which relies solely on image information for training, deliberately excluding depth data to maintain consistency with their approach.
The results, detailed in Table \ref{tbl:ft}, clearly demonstrate that our methods consistently outperform both traditional methods and FoodLMM across all evaluated tasks. This consistent superiority in performance underscores the effectiveness of our approach in tackling diverse challenges within the food domain.

\begin{figure}[] 
\centering
\includegraphics[width=1\linewidth, trim={0cm 0cm 0cm 0.5cm}, clip]{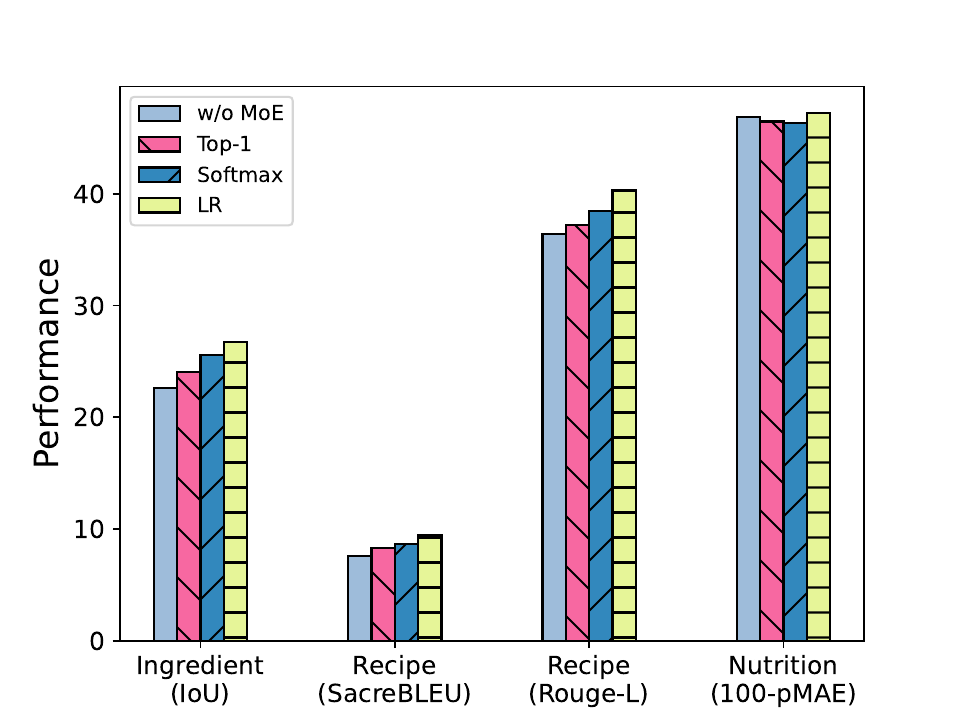}
\caption{Performance Comparison of Routing Strategies:  “Top-1” selects the top-1 expert accorarding router's score. “Softmax” applies a softmax operation to process the router's output, while “LR” uses linear rectified router. All the three strategies utilize a consistent LoRA rank setting of [8,8,8,8].
} 
\label{img:routing_strategy}
\end{figure}

\begin{figure*}[t] 
\centering
\includegraphics[width=1\linewidth]{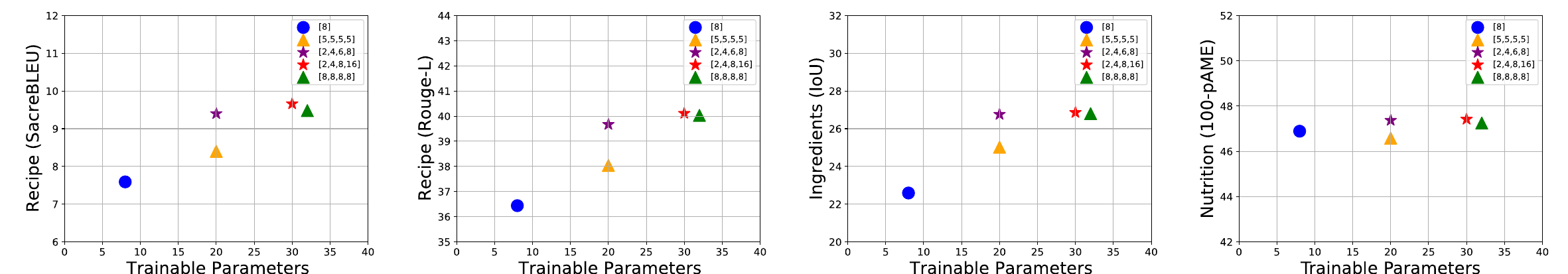}
\caption{\textbf{Correlation between the Quantity of Trainable LoRA Parameters and Task Performance Across Five LoRA Settings.} For simplicity, the total LoRA ranks in a single layer are quantified as the trainable parameters.
} 
\label{img:para_eff}
\end{figure*}

\begin{table*}[]
\centering
\caption{\textbf{Ablation Study on the Rank Settings of Experts within MoE.} The heterogeneous combination of experts yields superior performance while utilizing a reduced parameter count.}
\label{tbl:abla-rank}
% \resizebox{\textwidth}{!}{%
\begin{tabular}{l|c|c|cccc}
\toprule
Settings   & LoRA rank     & Routing strategy  & SacreBLEU ($\uparrow$ ) & Rouge-L ($\uparrow$ ) & Ingredient IoU ($\uparrow$ ) & pMAE ($\downarrow$ ) \\
\midrule
Var1  & 5, 5, 5, 5  &   LR  &  8.4   &  38.04   & 25.02  &   53.42       \\
Var2  & 8, 8, 8, 8 &   LR  &  9.48   &  40.03   & 26.7  &   52.75       \\
Var3  & 2, 4, 6, 8 &  LR  & 9.4   &  39.67   &   26.76    &    52.63       \\
Var4  & 2, 4, 8, 16 &  LR    & \textbf{9.66}   &  \textbf{40.11}   &   \textbf{26.86}    &    \textbf{52.58}       \\
\bottomrule
\end{tabular}
% }
\end{table*}

\subsection{Ablation study}
In this section, we perform an ablation study to evaluate the contribution and efficacy of the core components within our framework.
For simplicity, we omit the task names and tag only the metrics.

\subsubsection{MoE \textit{vs.} Larger LoRA rank}
As illustrated in Table \ref{tbl:abla-moe}, we first compare the standard LoRA (Var1) with its MoE variant (Var3). The results indicate a superior performance of the MoE design.
To ascertain whether this performance improvement can be attributed to an increase in the number of trainable LoRA parameters, we augment the LoRA rank of \textit{Var1} to 32 to obtain \textit{Var2}. 
A comparison between \textit{Var2} and \textit{Var4} reveals the superiority of the MoE design, even though these two variants have an equivalent number of trainable LoRA parameters. This finding suggests that the successful integration of MoE is not simply a result of expanding the number of trainable parameters.

\subsubsection{Routing Strategy}

Then we examine the effects of the routing strategy.
Our experiment includes three routing strategies: 1) similar to \cite{chen2024llavamole}, the top-1 expert is selected, referred to as \textbf{Top-1}; 2) following the approach in \cite{dou2023loramoe}, we employ a softmax function to normalize the router outputs, which is denoted as \textbf{Softmax}; 3) we introduce our novel linear rectified router, which we denote with the abbreviation \textbf{LR}. The results are summarized in Table \ref{tbl:abla-router} and visualized in Figure \ref{img:routing_strategy}.
As observed, while the Top-1 allocation MoE outperforms the standard LoRA (\textit{Var1}), it is surpassed by the "soft" allocation strategies (\ie \textit{Var2} and \textit{Var3}). This indicates the efficiency of employing an ensemble of experts to address a single task.
Moreover, the comparison between the LR and Softmax routers (\ie \textit{ Var2} and \textit{Var3}, \textit{Var4} and \textit{Var5}) suggests that the LR routing strategy is superior. These results further underscore the effectiveness of our linear rectification routing strategy.

\begin{figure}[h!]
    \centering
    \includegraphics[width=0.5\textwidth, trim={1cm 0.5cm 0.5cm 0.5cm}]{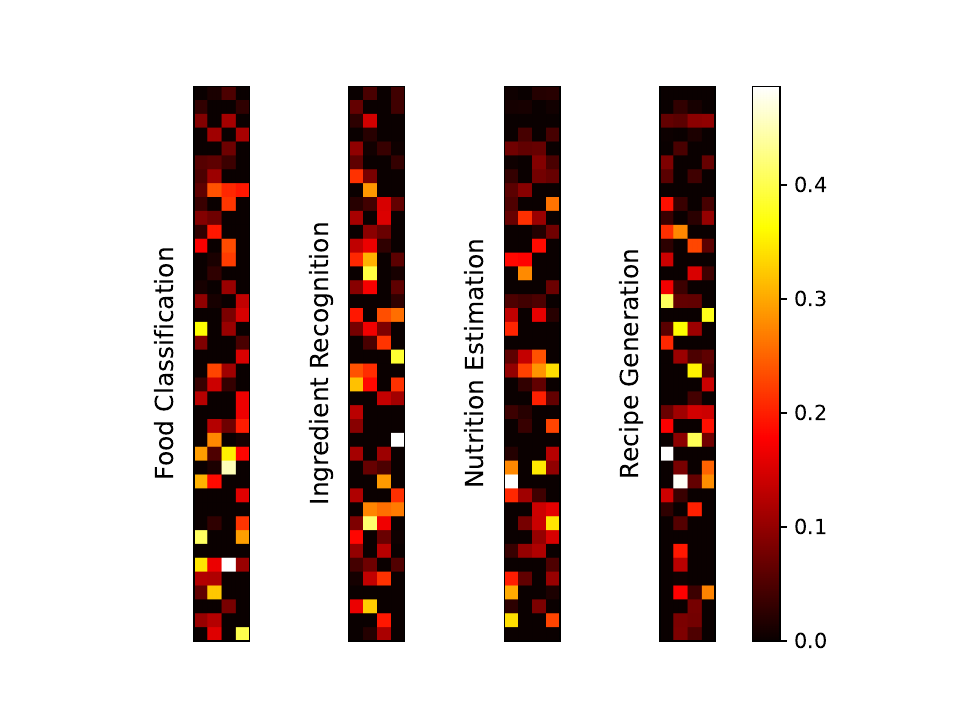}
    \caption{\textbf{Router Allocation Heatmaps for Different Tasks.} The horizontal axis represents the router outputs, while the vertical axis represents the top 40 transformer blocks.}
    \label{fig:heatmap_overall1}
\end{figure}

\textbf{Heatmap visualization of expert activation.} In addition, we visualize the heatmaps of router outputs for both the Softmax and LR routing strategies across several middle transformer blocks. Figure \ref{img:sparsity} presents this comparative analysis, clearly demonstrating that our proposed Linear Rectified Router achieves a higher degree of sparsity in task allocation.
Furthermore, we visualize the heatmaps of expert activation for the middle layer of top 40 transformer blocks across different tasks, as shown in Figure \ref{fig:heatmap_overall1}. The results indicate that different tasks engage different experts, with some tasks potentially sharing the same experts. These observations confirm that the RoDE model employs skill modules in a sparse manner and orchestrates skill sets in task-specific combinations, further substantiating the efficacy of our proposed RoDE approach.

\subsubsection{Rank configuration of LoRA}

We conduct an ablation study to evaluate the impact of different LoRA rank configurations. The results, as presented in Table \ref{tbl:abla-rank}, span multiple expert configurations, including:
1) \textit{Var1} - Four LoRAs, each with a uniform rank of 5.
2) \textit{Var2} - Four LoRAs, each with a uniform rank of 8.
3) \textit{Var3} - A heterogeneous set of four LoRAs with ranks of [2, 4, 6, 7].
4) \textit{Var4} - Another diverse set of four LoRAs with a distinct rank composition from \textit{Var3}, specifically [2, 4, 8, 16].
The analysis of \textit{Var1} and \textit{Var2} suggests that configurations with higher-ranked LoRA experts are more effective. However, despite having the same number of trainable parameters as \textit{Var1}, \textit{Var3} demonstrates superior performance, indicating that a mix of higher and lower-ranked LoRA experts can also be effective.
Moreover, \textit{Var3} shows comparable performance to \textit{Var2}, with only a slight reduction in SacreBLEU score (by 0.08) and Rouge-L score (by 0.36), while having 32.5\% fewer trainable parameters. This finding supports the notion that not all LoRA experts require high capacity to contribute effectively.

By adjusting the ranks in \textit{Var3} to create \textit{Var4}, we observe improved performance which exceeds that of \textit{Var2}, despite \textit{Var4} having 6.25\% fewer trainable parameters. This outcome implies that a strategic combination of LoRA expert sizes can effectively support the inclusion of some larger LoRAs, potentially enhancing their capability to handle complex tasks more efficiently.

\subsection{Additional Analysis}

\subsubsection{Training Parameter Efficiency}
We evaluate various rank settings by visualizing the correlation between the number of trainable LoRA parameters and their corresponding task performance on the Uni-Food dataset. Figure \ref{img:para_eff} illustrates this relationship, with the y-axis indicating task performance and the x-axis representing the trainable LoRA parameter count. 
It is apparent that the MoE architecture incorporates more trainable LoRA parameters, which substantially boosts performance across a range of metrics, \ie recipe SacreBLEU, recipe Rouge-L, and ingredient IoU. 
When comparing sets of homogeneous experts to those of heterogeneous experts (\ie comparing rank sets [5,5,5,5] to [2,4,6,8]), the introduction of high rank LoRA can enhance model performance despite when the number of trainable LoRA parameters is the same.
These results suggest that not every expert requires an extensive parameter set; heterogeneous experts can effectively coordinate trainable parameters, resulting in better performance on multi-task learning.

\subsubsection{Empirical Analysis}
\begin{figure*}[t]
    \centering
    \includegraphics[width=1\textwidth]{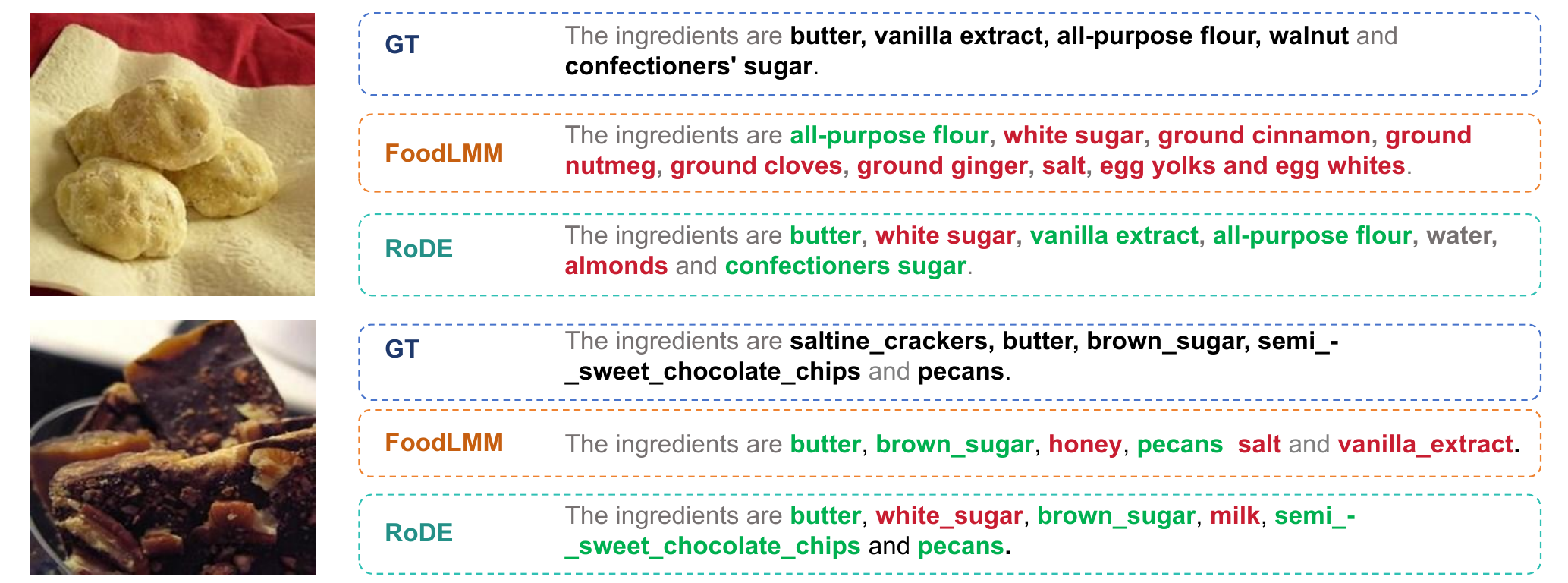}
    \caption{Comparison between FoodLMM and RoDE on ingredient recognition.}
    \label{fig:cp_ir_1}
\end{figure*}

We present some examples of ingredient recognition on FoodLMM and RoDE, as depicted in Figure \ref{fig:cp_ir_1}. In these examples, RoDE demonstrates higher accuracy in ingredient recognition. However, it still incorrectly identifies some ingredients, and tends to predict commonly used but ambiguous ingredients.

\subsubsection{Computational Overhead}
RoDE employs the rectified linear union, a strategy tailored for sparse expert activation, which is efficient for both training and inference. The average training time for a single batch and the average inference time for recipe generation on the UniFood dataset are demonstrated in Table \ref{tbl:time}. 
Remarkably, while RoDE exhibits superior performance compared to the standard MoE design (as shown in Table ~\ref{tbl:uni-food} and Table ~\ref{tbl:abla-router}), it manages to maintain a time consumption that is comparable to that of the standard MoE design.

\begin{table}[h!]
\centering
\caption{\textbf{Training and inference computation cost.}  The training procedure utilizes a batch size of 4 and is conducted on four RTX4090 GPUs. While the inference process is carried out on a single RTX4090 GPU.}
\label{tbl:time}
\resizebox{0.48\textwidth}{!}{%
\begin{tabular}{lccc}
\toprule
   & FoodLMM & FoodLMM+MoE & RoDE  \\
\midrule
Training per batch & 21.87s   & 23.68s       & 23.72s \\
Inference on recipe generation & 19.21s & 20.41s & 20.31s \\
\bottomrule
\end{tabular}
}
\end{table}

\section{Conclusion}
In this paper, we delve into the broader scope of tasks within the realm of food studies. We introduce Uni-Food, a comprehensive dataset comprising classification, ingredient recognition, recipe generation, and nutrition estimation. Uni-Food serves as a foundational resource empowering a wide spectrum of food-related research endeavors.
The expansion of training tasks brings about a challenge of task conflict.
To mitigate these task conflicts during SFT on Large LMMs, we introduce the Linear Rectified Mixture of Diverse Experts (RoDE). 
RoDE constructs a skill space that experts are sharable, and arranges a variety of trainable parameters to establish heterogeneous experts capable of adapting to tasks of different complexities.
It employs linear rectified units to refine linear routers, encouraging the routers to learn sparse allocation.
RoDE is not only GPU memory-efficient but it also simplifies the optimization process. Our experimental results clearly demonstrate the effectiveness of our proposed method, highlighting its efficiency and efficacy in overcoming the multifaceted challenges associated with the food domain.

\bibliographystyle{IEEEtran}
\small\bibliography{egbib}

\end{document}

\title{RoDE: Linear Rectified Mixture of Diverse Experts for Food Large Multi-Modal Models}

\author{Pengkun~Jiao,
        Xinlan~Wu,
        Bin~Zhu,
        Jingjing~Chen,
        Chong-Wah~Ngo and Yu-Gang~Jiang % <-this % stops a space

% \thanks{Pengkun Jiao is with the School of Computer Science, Fudan University. 
% Shanghai, China. E-mail: pkjiao21@m.fudan.edu.cn. }
% \thanks{Xinlan Wu is with the School of Computer Science, Fudan University. 
% Shanghai, China. E-mail: pkjiao21@m.fudan.edu.cn. }
% \thanks{Bin Zhu is with the School of Computing and Information Systems, Singapore Management University. 
% Shanghai, China. E-mail: pkjiao21@m.fudan.edu.cn. }
% \thanks{Jingjing Chen is with the School of Computer Science, Fudan University. 
% Shanghai, China. E-mail: chenjingjing@fudan.edu.cn. }
% \thanks{Chong-Wah Ngo is with the School of Computing and Information Systems, Singapore Management University.}
% \thanks{Yu-gang Jiang is with the School of Computer Science, Fudan University. 
% Shanghai, China. E-mail: ygj@fudan.edu.cn }
}

\maketitle

\renewcommand{\thesection}{\Alph{section}}

\section{prompting strategy}

\begin{figure*}[h!]
    \centering
    \includegraphics[width=0.95\textwidth]{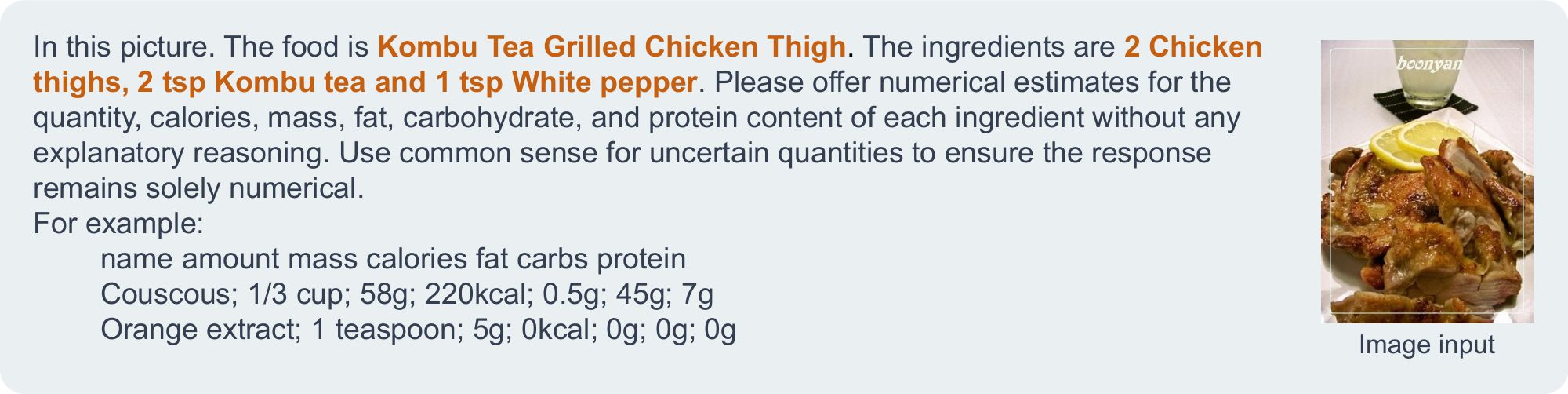}
    \caption{Example prompt for nutrition annotation collection from ChatGPT4v. The input prompt includes text input and image input.}
    \label{fig:prompt}
\end{figure*}

We input the food image along with its quantified ingredient list into ChatGPT-4V to obtain nutritional information. An illustrative example of the prompt design is depicted in Figure~\ref{fig:prompt}.

\section{Nutrition Information Compared with Recipe1M+}
\begin{figure*}[]
\centering
\includegraphics[width=1\linewidth]{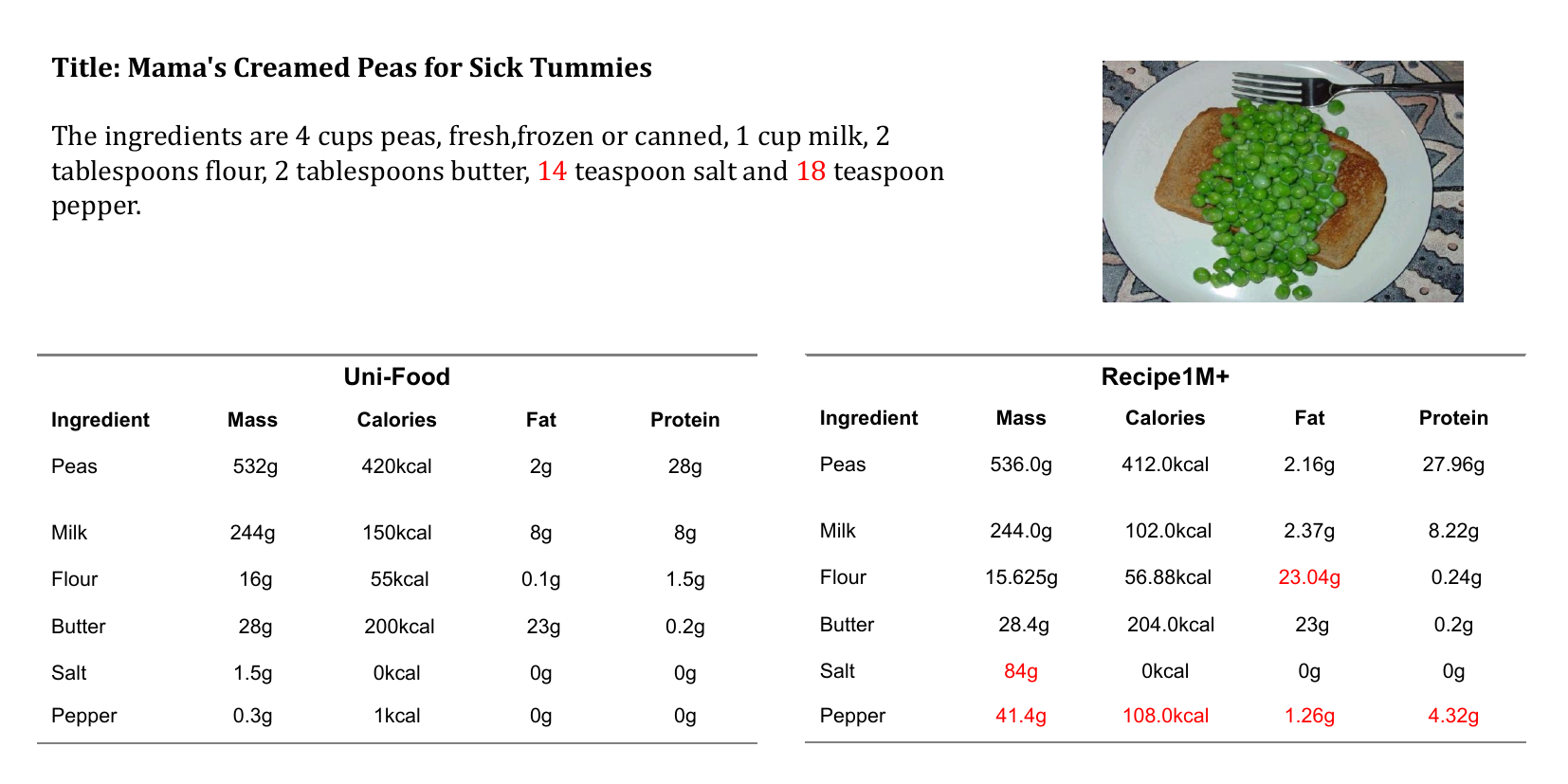}
\caption{Nutrition information comparison between Uni-Food and Recipe1M+. The table lookup strategy adopted by Recipe1M+ becomes uncertain when the annotation of the source ingredient quantity is incorrect. The ingredient annotations are sourced from Recipe1M.
}
\label{img:nut_compare2}
\end{figure*}

\begin{figure*}[]
\centering
\includegraphics[width=1\linewidth]{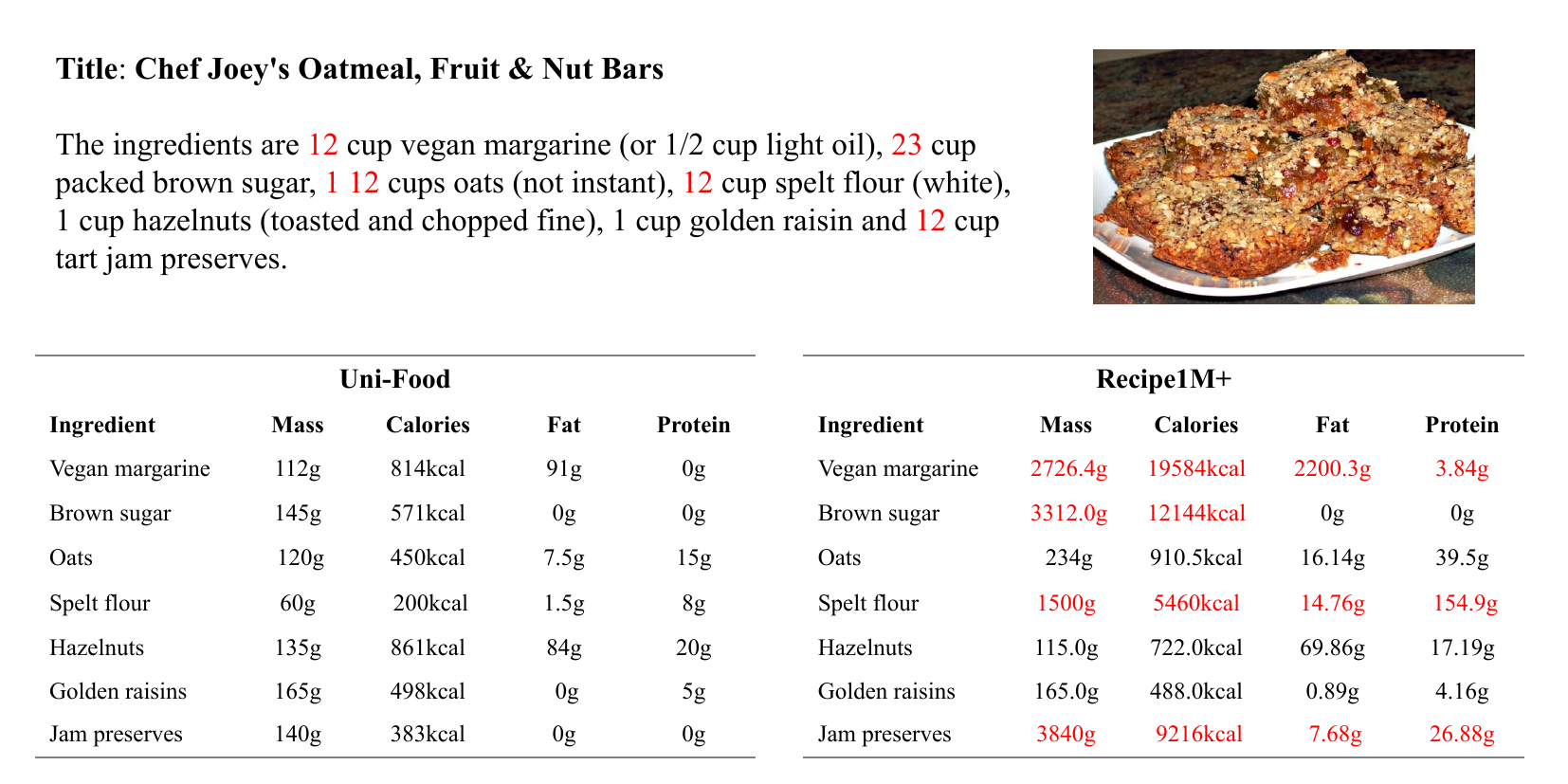}
\caption{Nutrition information comparison between Uni-Food and Recipe1M+.
}
\label{img:nut_compare3}
\end{figure*}

\begin{figure*}[]
\centering
\includegraphics[width=1\linewidth]{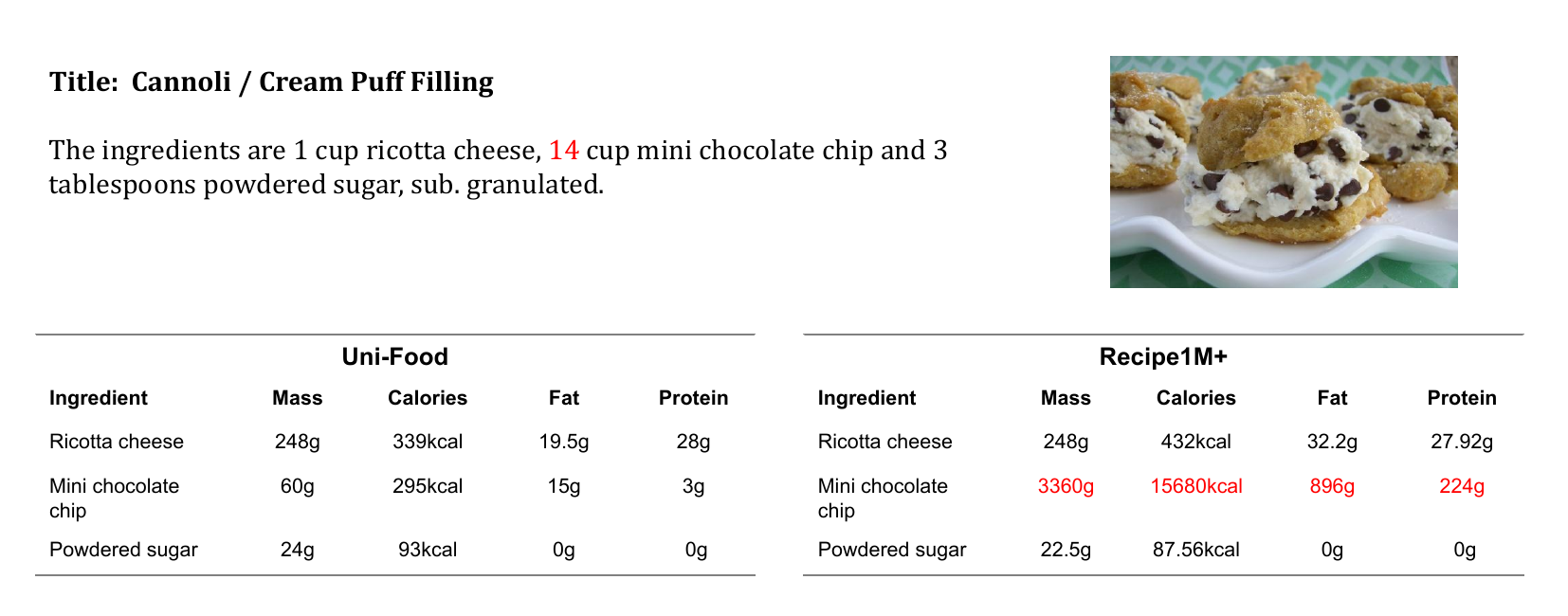}
\caption{Nutrition information comparison between Uni-Food and Recipe1M+.
}
\label{img:nut_compare4}
\end{figure*}

\begin{figure*}[]
\centering
\includegraphics[width=1\linewidth]{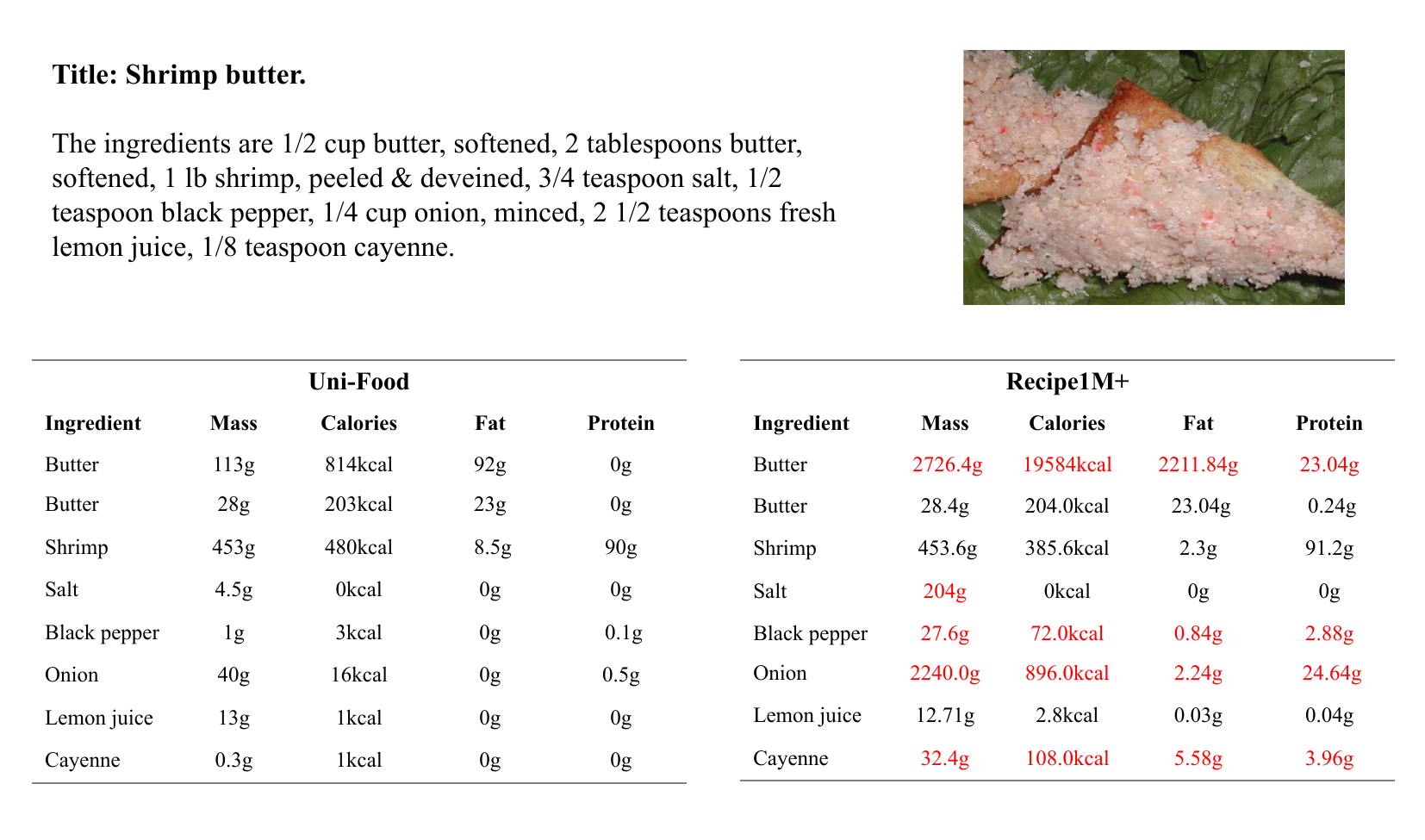}
\caption{Nutrition information comparison between Uni-Food and Recipe1M+. The latter may contain abnormal nutritional information even when the source ingredient quantity is correct.
}
\label{img:nut_compare1}
\end{figure*}

We meticulously examined the nutritional information of ingredients within Recipe1M+ \cite{marin2019recipe1m} and uncovered certain uncertainties. 
Notably, Recipe1M+ employs a table lookup strategy to extract nutritional information from the USDA\footnote{https://fdc.nal.usda.gov/} database, relying heavily on the accurate annotation of ingredient quantities. However, we discovered instances where ingredient annotations were incorrect. As depicted in Figures \ref{img:nut_compare2}, \ref{img:nut_compare3}, and \ref{img:nut_compare4}, several samples exhibited abnormal quantities of certain ingredients. For instance, in Figure \ref{img:nut_compare2}, the pea food recipe indicated 14 teaspoons of salt and 18 teaspoons of pepper, which defies conventional measures; the correct amounts should be 1/4 teaspoon of salt and 1/8 teaspoons of pepper. Consequently, Recipe1M+ translates these inaccuracies into mass, resulting in excessively large quantities.
Moreover, even in some cases where ingredients are correctly annotated, Recipe1M+ may still exhibit uncertainties, as evidenced in Figure \ref{img:nut_compare1}. While ingredient quantities are accurately annotated, Recipe1M+ presents abnormal nutritional information.
In contrast, our Uni-Food dataset harnesses the capabilities of the robust LMM, ChatGPT-4Vision\cite{achiam2023gpt4}, which utilizes image information robust to ingredient quantity annotation noise and proficiently interprets units, resulting in precise nutritional estimates.

\section{Additional Illustrations of Expert Activation Across Different Tasks}

\begin{figure*}[t]
    \centering
    \includegraphics[width=0.8\textwidth]{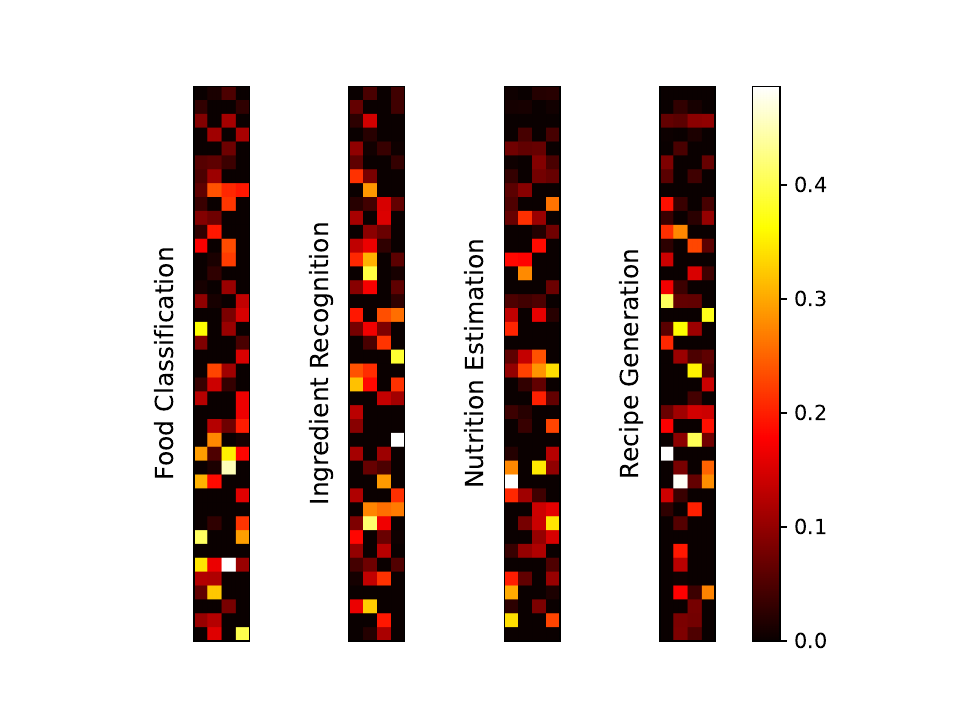}
    \caption{\textbf{Router Allocation Heatmaps for Different Tasks.} The horizontal axis represents the router outputs, while the vertical axis represents the top 40 transformer blocks.}
    \label{fig:heatmap_overall2}
\end{figure*}

\begin{figure*}[t]
    \centering
    \includegraphics[width=0.8\textwidth]{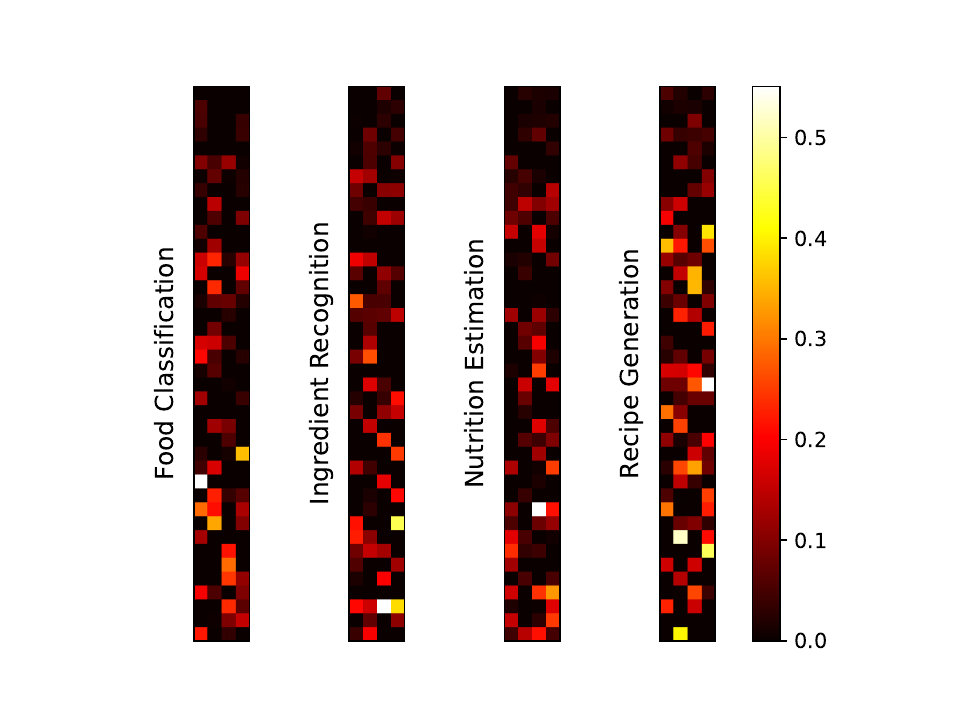}
    \caption{\textbf{Router Allocation Heatmaps for Different Tasks.} The horizontal axis represents the router outputs, while the vertical axis represents the top 40 transformer blocks.}
    \label{fig:heatmap_overall3}
\end{figure*}

We select two additional images and display the expert activation heatmaps of the adaptive modules from the transformer blocks, as illustrated in Figures \ref{fig:heatmap_overall2} and \ref{fig:heatmap_overall3}.

% \section{Empirical Analysis}
% We present some examples of ingredient recognition on FoodLMM and RoDE, as depicted in Figure \ref{fig:cp_ir_1}.

% \begin{figure*}[h!]
%     \centering
%     \includegraphics[width=1\textwidth]{imgs/exa_ana.pdf}
%     \caption{Comparison between FoodLMM and RoDE on ingredient recognition.}
%     \label{fig:cp_ir_1}
% \end{figure*}

\bibliographystyle{IEEEtran}
\small\bibliography{egbib}